\documentclass[10pt,journal,compsoc]{IEEEtran}

\ifCLASSOPTIONcompsoc

\usepackage[nocompress]{cite}

\else
\usepackage{cite}
\fi
\ifCLASSINFOpdf
\else
\fi
\usepackage{amsthm}
\usepackage{amsmath}
\usepackage{graphics}
\usepackage{graphicx}
\usepackage{txfonts}
\usepackage{subfigure}
\usepackage[level]{datetime} 
\usepackage{algorithm}
\usepackage{algorithmic}
\usepackage{caption}
\usepackage{float}
\usepackage{threeparttable}
\usepackage{url}
\usepackage{multirow}
\usepackage{diagbox}
\usepackage{hyperref}
\newtheorem{Lemma}{Lemma}
\newtheorem{theorem}{Theorem}
\newtheorem{corollary}{Corollary}

\newtheorem{proposition}{Proposition}

\begin{document}
\title{ELBD: Efficient score algorithm for feature selection on latent variables of VAE}
\author{Yiran Dong   and Chuanhou~Gao,~\IEEEmembership{Senior Member,~IEEE}
\IEEEcompsocitemizethanks{\IEEEcompsocthanksitem Y. Dong and C. Gao are with the School of Mathematical Sciences, Zhejiang University, Hangzhou 310030, China.\protect\\
E-mail:  \{22035082, gaochou\}@zju.edu.cn
\IEEEcompsocthanksitem }
\thanks{Manuscript received \date{}.}}
%\markboth{IEEE Transactions on Pattern Analysis and Machine Intelligence}
%{Shell \MakeLowercase{\textit{et al.}}: Bare Demo of IEEEtran.cls for Computer Society Journals}

\IEEEtitleabstractindextext{
\begin{abstract}
In this paper, we develop the notion of evidence lower bound difference (ELBD), based on which an efficient score algorithm is presented to implement feature selection on latent variables of VAE and its variants. Further, we propose weak convergence approximation algorithms to optimize VAE related models through weighing the ``more important" latent variables selected and accordingly increasing evidence lower bound. We discuss two kinds of different Gaussian posteriors, mean-filed and full-covariance, for latent variables, and make corresponding theoretical analyses to support the effectiveness of algorithms. A great deal of comparative experiments are carried out between our algorithms and other 9 feature selection methods on 7 public datasets to address generative tasks. The results provide the experimental evidence of effectiveness of our algorithms. Finally, we extend ELBD to its generalized version, and apply the latter to tackling classification tasks of 5 new public datasets with satisfactory experimental results.

%This paper proposes weak convergence approximation algorithms to increase the evidence lower bound of VAE. It uses selected latent variables to optimize VAE models and its variants including state-of-the-art generative model, and we provide bunch of theorems to guarantee its effectiveness. To get selected latent variables which are used in weak convergence approximation algorithms, feature selection algorithms are needed. We propose evidence lower bound difference(ELBD) score for this purpose specifically. We also use other frequently-used feature selection algorithms to select latent variables and compare them with ELBD score. To extend the application of ELBD, we also propose generalized evidence lower bound difference(gELBD) score and compare it with other frequently-used algorithms on classification tasks

\end{abstract}

}

\maketitle
\IEEEdisplaynontitleabstractindextext
\IEEEpeerreviewmaketitle

\begin{figure}[htbp]
\centering
\includegraphics[scale=0.4]{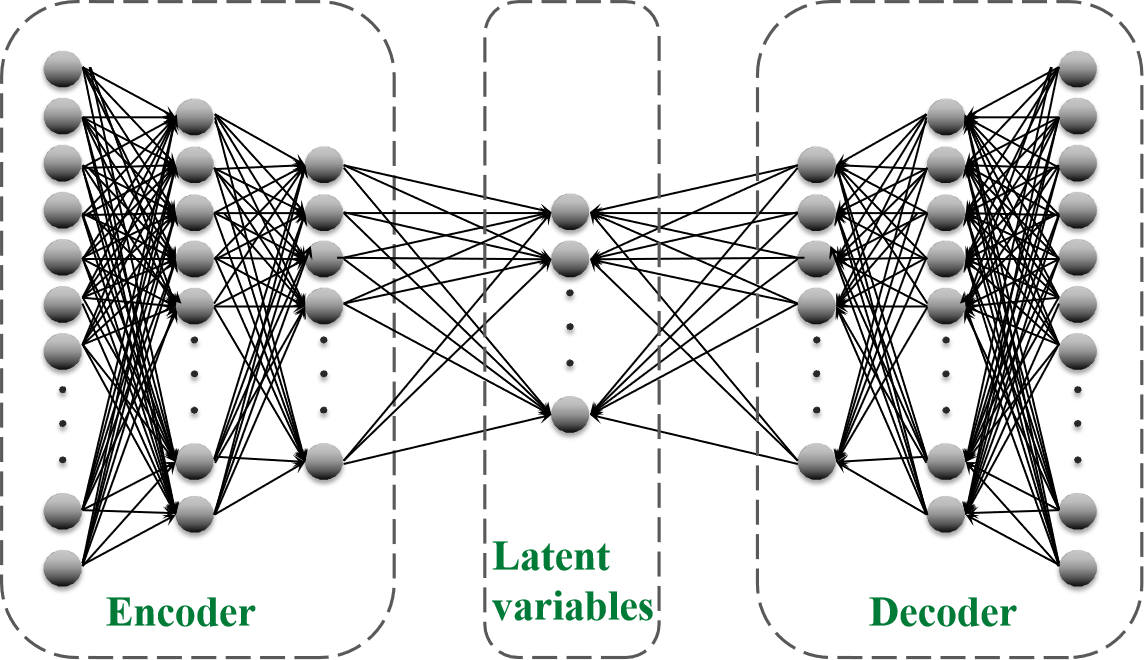}
\caption{VAE with 3 layers of encoder and 3 layers of decoder.}\label{fig1:vae}
\end{figure}

\IEEEraisesectionheading{\section{Introduction}}
\label{intro}
\IEEEPARstart{V}{ariational} autoencoder (VAE) has been one of the most popular modeling methods since it was pioneered by Kingma and Welling \cite{kingma2013auto}. Essentially, it is a kind of generative modeling approach that handles models of distribution about data points. VAE is built on neural networks and trained with stochastic gradient descent. From the viewpoint of structure, there includes encoder, latent variables layer and decoder in a VAE model, and for encoder and decoder there may be multi-layers of hidden layer therein. Figure \ref{fig1:vae} exhibits a schematic of VAE. Throughout the working process, the encoder encodes the original samples into latent variables while the decoder reconstructs new samples from latent variables. The whole workflow is implemented through unsupervised learning of neural networks. Benefiting from complete mathematical support, VAE has exhibited strong power in generating many complex data, such as
handwritten digits \cite{kingma2013auto} and human faces \cite{doersch2016tutorial}.

Following the rapid development of VAE, some improved versions consecutively emerge. Conditional variational autoencoder (CVAE) \cite{sohn2015learning}, instead of unsupervised learning in VAE, models  complicated distributions about data points through supervised learning. Delta variational autoencoder ($\delta$-VAE)\cite{razavi2019preventing} uses first-order linear autoregressive process on latent variables to get new latent variables which can improve the flexibility of posterior distribution. Inverse autoregressive transformations VAE (IAF-VAE)\cite{kingma2016improved} uses neural network to do the autoregression on latent variables and can improve the generative ability dramatically. Vector quantisation VAE (VQ-VAE)\cite{razavi2019generating} discretizes latent variables such that they fit many of the modalities in nature. Mutual posterior-divergence regularization for VAE (MAE)\cite{ma2019mae} changes the $\mathcal{KL}$ divergence in the loss of VAE as mutual posterior diversity. Nouveau VAE (NVAE)\cite{vahdat2020nvae} focuses on the network structure of VAE and makes VAE to be able to generate much more large pictures clearly.

VAE and its variants have been used in many areas. In chemistry and biology, VAE can discover and generate molecular graphs which lowers the cost of drug design and accelerates the whole process of drug producing\cite{samanta2020nevae}. In natural language processing, VAE can be trained on text and help to interpolate or complete the sentences\cite{bowman2015generating}. By projecting the data set into latent spaces whose dimension is much smaller than dimension of original data, VAE can compress the data set\cite{lombardo2019deep}. VAE will use decoder to recover the whole data set efficiently and accurately if we want. After encoding the pictures as latent variables, we can adjust the latent variables in a particular way and feed them to decoder to get synthetic pictures we want. This technique is often used in image synthesis area\cite{white2016sampling}.

 %Generative models are widely used in machine learning. In reinforcement learning, generative models can be used to simulate possible future or help agent make decisions.\cite{} In experiments, we may face the incomplete data, generative models can make probabilities on every possible values of missing features, and take the probabilities as weights for this complete data. Furthermore, we also can turn the generative models as discriminators or classifier by using Bayes rule.

We note that nearly all the improvements on VAE are strongly related to latent variables. Latent variables show the importance even in some industrial applications. We propose the weak convergence approximation algorithms which use part of important latent variables to improve the generative ability of VAE and its variants. The principle of the algorithms is that they equivalently increase the weight of distributions of part of latent variables. Hence we need feature selection algorithms to select latent variables.

Feature selection is widely used during data processing. Feature selection algorithms score every feature and pick the features according to the scores from the largest to the smallest\cite{li2017feature}. The aim of them is to pick features which contain useful information and help model avoid overfitting and dimension explosion problems\cite{kumar2014feature}. Naturally, it is not a novel project in the field of machine learning, and some classical methods, such as Laplacian score\cite{hastie2009elements}, relevant feature selection\cite{robnik2003theoretical} etc., have been widely used in data processing. Here, we continue to use this strategy to select latent variables of VAE. Moreover, we propose evidence lower bound difference (ELBD) score to do the selection on latent variables. ELBD uses absolute value of difference of two evidence lower bounds (ELBOs) to score each latent variable. Based on ELBD, generalized ELBD (gELBD) is proposed, and we find that gELBD performs well on classification tasks. We compare ELBD and gELBD with other feature selection algorithms including state-of-the-art algorithms in experiments.

% However, unlike those traditional dimension reduction techniques where various projections usually need to be defined between original features and the reduced ones, the current proposed method (called adaptive dimension reduction algorithm in the context), getting inspiration from sequential minimal optimization \cite{Platt1998SequentialMO}, works through judging whether each feature contains too much noise. We put forward a new pattern to realize the above process. By introducing $\boldsymbol \pi$ vector whose elements are 0 or 1, and further utilizing the element-wise multiplication between $\boldsymbol \pi$ and latent variables, the algorithm will remove those latent variables inactivated by 0 component of $\boldsymbol\pi$ while keep those activated by 1 component of $\boldsymbol\pi$. The algorithm is also proved to have convergence and local convexity about the output. These properties are further verified by a great deal of numerical experiments. The potential effect of the algorithm is, on the one hand, to obtain higher precision and more efficient VAE related models, one the other hand, to generate a complete new high-efficiency model. 

The rest of the paper is organized as follows. Section 2 presents some preliminaries about VAE and its variants, for which two kinds of latent variables are used and the algorithms of training process are proposed, respectively. In Section 3, we state our motivation and purpose. Then we propose the weak convergence approximation algorithms on different types of latent variables, and give theoretical analyses about them. In Section 4, we propose our own feature selection algorithm and generalize it. This is followed by some experimental studies of proposed algorithms in Section 5. Finally, Section 6 concludes the paper. 

\textbf{Notation:} Throughout the paper, the bold lowercase letters and Greeks represent row vectors, bold capital letters and Greeks represent matrices, and $(\mathbf z)_i$ or $z_i$ represents the $i$th element of row vector $\mathbf z$. We also define the norm on vector $\Vert\mathbf z\Vert=\underset{i}{\max}|z_i|$. In addition, only pictures data sets are considered.

\section{Preliminaries}
In this section, we will introduce shortly VAE and its variants\cite{doersch2016tutorial}, then give the algorithms of training processes for VAE with two kinds of different latent variables.

\subsection{VAE}
VAE is a kind of typical latent variable model, whose main purpose is to generate sample points sharing the same distribution with the original samples as far as possible through utilizing latent variables. 

Let $\mathbb{D}=\{\mathbf x_1,...,\mathbf x_N\}$ represents a data set, and the task is to infer the distribution of $\mathbf{x}$, $P(\mathbf x)$, so as to generate new sample points. VAE addresses this issue by introducing latent variables $\mathbf{z}\in\mathbb{R}^m$ obeying the standardized normal distribution $\mathcal{N}(\mathbf 0,\mathbf I)$, and further minimizing the $\mathcal{KL}$ divergence between the true posterior distribution $P(\mathbf z|\mathbf x)$ and the learned one $Q(\mathbf z|\mathbf x)$ on $\mathbb{D}$. Here, the $\mathcal{KL}$ divergence is essential about the expectation calculation, defined by
\begin{align}\label{loss1}
\mathcal{KL}[Q(\mathbf z|\mathbf x)\parallel P(\mathbf z|\mathbf x)]=E_{Q(\mathbf z|\mathbf x)}\left(\log Q(\mathbf z|\mathbf x)-\log P(\mathbf z|\mathbf x)\right).
\end{align} 
By adding $P(\mathbf x)$ and applying Bayes rule to $P(\mathbf z|\mathbf x)$, the above expression can be rewritten as
\begin{align}
&\displaystyle\log P(\mathbf x)-\mathcal{KL}[Q(\mathbf z|\mathbf x)\parallel P(\mathbf z|\mathbf x)]\notag\\
=&E_{Q(\mathbf z|\mathbf x)}\left(\log P(\mathbf x|\mathbf z)\right)-\mathcal{KL}[Q(\mathbf z|\mathbf x)\parallel P(\mathbf z)],\label{ELBO}
\end{align} 
which is the core formula in VAE. The left hand side of the formula is just right the maximization objective with the first term to represent the log-likelihood of $P(\mathbf x)$ while the second term to express minus error of estimating $P(\mathbf z|\mathbf x)$ by $Q(\mathbf z|\mathbf x)$. The right hand side takes a form of autoencoder. The whole forward process contains following steps: $Q(\mathbf z|\mathbf x)$ which is called encoder, ``encodings" original data $\mathbf x$ into $\mathbf z$, and the ``reparameterization trick" is adopted on $\mathbf z$. Then $P(\mathbf x|\mathbf z)$ which is called decoder, ``decodings" it to reconstruct $\mathbf x$. The right side of equation (\ref{ELBO}) can be also seen as a lower bound of $\log P(\mathbf x)$ due to the nonnegativity of $\mathcal{KL}[Q(\mathbf z|\mathbf x)\parallel P(\mathbf z|\mathbf x)]$, usually called evidence lower bound (ELBO). The inner two terms are both computable, so the negative ELBO often acts as the cost function one can be optimized via stochastic gradient descent.

To optimize equation (\ref{ELBO}), it needs to know a specific form about $Q(\mathbf z|\mathbf x)$. The usual choice for it, and also for $P(\mathbf x|\mathbf z)$ and $P(\mathbf z)$ is the Gaussian distribution, i.e.,
\begin{align}\label{distribution}
Q(\mathbf z|\mathbf x)=\mathcal N(\boldsymbol\mu(\mathbf x),\boldsymbol\Sigma(\mathbf x)), ~P(\mathbf x|\mathbf z)=\mathcal N(\mathbf f(\mathbf z),\mathbf I), ~ p(\mathbf z)=\mathcal N(0,\mathbf I),
\end{align} 
where $\boldsymbol\mu$, $\boldsymbol\Sigma$ and $\mathbf f$ are arbitrary deterministic functions that can be learned via neural networks. $\mathbf I$ is identity matrix. With these distributions, the back propagation can compute a gradient used for stochastic gradient descent, which completes the learning process.  

CVAE is a kind of conditional VAE that conditions on the labels of data $\mathbf y$ during the entire generative process. Encoder and decoder become $Q(\mathbf z|\mathbf x,\mathbf y)$ and $P(\mathbf z|\mathbf x,\mathbf y)$. So it is more potential in handling images that have different categories.

Norm flow VAE (NF-VAE)\cite{rezende2015variational} uses autoregressive function $\mathbf z_t=\mathbf z_{t-1}+\mathbf ph(\mathbf c^\top\mathbf z_{t-1}+b)$ to improve the flexible of posterior distribution of latent variables where $h$ is a invertible function and the common choice of $h$ is $\tanh$. NF-VAE is also the stronger version of $\delta$-VAE which treats $h$ as linear function. After going through $T$ norm flowing layers, the latent variables' distribution can be computed by the equation 
\begin{align*}
    \log Q(\mathbf z_T|\mathbf x)=\log Q(\mathbf z_0|\mathbf x)-\sum^T_{t=1}\log\det\bigg |\frac{d\mathbf z_t}{d\mathbf z_{t-1}}\bigg|.
\end{align*}

Inverse autoregressive flow VAE (IAF-VAE) is a modification of NF-VAE and it is also one of the state-of-the-art generative models. Instead of using $h$, it uses the network to form the autoregressive layers.
\begin{eqnarray*}\label{loss3C}
~\left\{
\begin{array}{lll}
\mathbf m,\mathbf s=AutoregressiveNetwork(\mathbf z_{t-1},\mathbf h), \\
\mathbf c=sigmoid(\mathbf s),\\
\mathbf z_t=\mathbf c\otimes\mathbf z_{t-1}+(\mathbf 1-\mathbf c)\otimes\mathbf m.
\end{array}
\right.
\end{eqnarray*}
where the symbol $\otimes$ represents the element-wise multiplication and $\mathbf h$ is one of the output of encoder.

The aims of NF-VAE and IAF-VAE are that they expand the range of latent variables' distribution, make them not only be constrained in the family of Gaussian distribution. However, we can also see them as a part of decoder function in the forward process and we only consider $\mathbf z_0$ in NF-VAE and IAF-VAE whose posterior distribution is still Gaussian distribution in equation (\ref{distribution}).

In this and latter sections, we only show algorithms on original VAE, the corresponding algorithms on CVAE, NF-VAE, IAF-VAE or any other variant models of VAE can be expanded naturally. And in Section 5, we take experiments on VAE, CVAE, NF-VAE and IAF-VAE.

\subsection{Mean-field and full-covariance Gaussian posterior}
This subsection introduces two types of Gaussian posterior of latent variables: mean-field and full-covariance\cite{kingma2019introduction}.
The mean-field Gaussian posterior assumes the $\boldsymbol\Sigma(\mathbf x)$ in $Q(\mathbf z|\mathbf x)$ as the diagonal matrix. Oppositely, the full-covariance Gaussian posterior makes $\boldsymbol\Sigma(\mathbf x)$ be a full matrix to train.  We take original VAE as an example, and the training processes are shown in \textbf{Algorithm \ref{mean-field}} and \textbf{Algorithm \ref{full-covariance}}.

\begin{algorithm}
\renewcommand{\algorithmicrequire}{\textbf{Input:}}
\renewcommand{\algorithmicensure}{\textbf{Output:}}
\caption{The training process of VAE with mean-field Gaussian posterior}
\label{mean-field}
\begin{algorithmic}[1]
\REQUIRE data set $\mathbb D=\{\mathbf x_1,...,\mathbf x_N\}$
\FOR{ data $\mathbf x$ in $\mathbb D$}
\STATE $\boldsymbol\mu$, $\log\boldsymbol\sigma^2= EncoderNetwork(\mathbf x)$;
\STATE sample $\boldsymbol\epsilon$ from normal distribution $\mathcal N(0,\mathbf I)$;
\STATE $\mathbf z=\boldsymbol\epsilon\otimes\exp(\frac12\log\boldsymbol\sigma^2)+\boldsymbol\mu$;
\STATE $\mathbf x'=DecoderNetwork(\mathbf z)$;
\STATE $loss=\frac12\overset{\dim\mathbf x}{\underset{i=1}{\sum}} (\mathbf x'-\mathbf x)_i^2+\frac12\overset{\dim\mathbf z}{\underset{i=1}{\sum}} (\boldsymbol\mu^2+\exp(\log\boldsymbol\sigma^2)-\log\boldsymbol\sigma^2-1)_i$;
\STATE optimize the $loss$ by gradient descent methods;
\ENDFOR
\end{algorithmic}
\end{algorithm}

We put original data $\mathbf x$ in the encoder network and obtain the $\boldsymbol\mu$ and $\log\boldsymbol\sigma^2$ where $\boldsymbol\mu$ is the mean of $Q(\mathbf z|\mathbf x)$, $\boldsymbol\sigma^2$ is the diagonal of the covariance matrix $\boldsymbol\Sigma(\mathbf x)$ in equation (\ref{distribution}). Line 3 and Line 4 are called reparameterization. It generates latent variables $\mathbf z$ and makes sure $\mathbf z$ follow $\mathcal N\left(\boldsymbol\mu(\mathbf x),\boldsymbol\Sigma(\mathbf x)\right)$. Then we feed $\mathbf z$ in the decoder network to get a new data $\mathbf x'$ in line 5. The loss function in line 6 is negative ELBO after putting the equation (\ref{distribution}) into equation (\ref{ELBO}) where $\dim\mathbf x$ and $\dim\mathbf z$ are the dimension of row vectors $\mathbf x$ and $\mathbf z$. And it is also the loss we need to optimize in line 7.

\begin{algorithm}
\renewcommand{\algorithmicrequire}{\textbf{Input:}}
\renewcommand{\algorithmicensure}{\textbf{Output:}}
\caption{The training process of VAE with full-covariance Gaussian posterior}
\label{full-covariance}
\begin{algorithmic}[1]
\REQUIRE data set $\mathbb D=\{\mathbf x_1,...,\mathbf x_N\}$
\FOR{ data $\mathbf x$ in $\mathbb D$}
\STATE $\boldsymbol\mu$, $\log\boldsymbol\sigma$, $\mathbf L_m=EncoderNetwork(\mathbf x)$;
\STATE mask matrix $\mathbf L_m$ and turn the $\mathbf L_m$ to lower triangular matrix with zeros on and above the diagonal;
\STATE $\mathbf L=\mathbf L_m+diag(\exp(\boldsymbol\sigma))$;
\STATE sample $\boldsymbol\epsilon$ from normal distribution $\mathcal N(0,\mathbf I)$;
\STATE $\mathbf z=\boldsymbol\mu+\boldsymbol\epsilon\mathbf L$;
\STATE $\mathbf x'=DecoderNetwork(\mathbf z)$;
\STATE $L_1=-\frac12\overset{\dim\mathbf z}{\underset{i=1}{\sum}}(\boldsymbol\epsilon^2+\log(2\pi)+\log\boldsymbol\sigma)_i$;
\STATE $L_2=-\frac12\overset{\dim\mathbf z}{\underset{i=1}{\sum}}(\mathbf z^2+\log(2\pi))_i$;
\STATE $L_3=\frac12\overset{\dim\mathbf x}{\underset{i=1}{\sum}} (\mathbf x'-\mathbf x)_i^2$
\STATE $loss=L_3+L_1-L_2$;
\STATE optimize the $loss$ by gradient descent methods;
\ENDFOR
\end{algorithmic}
\end{algorithm}

Instead of outputting $\log\boldsymbol\sigma^2$, encoder network outputs $\log\boldsymbol\sigma$ and $\mathbf L_m$ in full-covariance condition. We still take $\boldsymbol\sigma^2$ as the diagon of $\boldsymbol\Sigma(\mathbf x)$. After the mask operation in line 3, we see the $\mathbf L$ in line 4 as the Cholesky factorization of symmetric positive definite matrix $\boldsymbol\Sigma(\mathbf x)$\cite{greub2012linear}. Then latent variables $\mathbf z$ follow $\mathcal N\left(\boldsymbol\mu,\mathbf L\mathbf L^\top\right)$ after reparameterization. It should be noted that in line 11, $L_1-L_2$ is not the exact $\mathcal{KL}$ divergence but the unbiased estimation of it, i.e., $E_{Q(\mathbf z|\mathbf x)}(L_1-L_2)=\mathcal{KL}[Q(\mathbf z|\mathbf x)\parallel P(\mathbf z)]$.

\section{Optimize VAE using selected features}

In this section, we state how we use the features we pick to increase the ELBO and why we do the features selection on latent variables.

\subsection{Marginalization of encoder}
For the sake of simplicity and clarity, we separate the latent variables $\mathbf z$ as two sets $\mathbf w$ and $\mathbf u$, i.e., $\mathbf z=(\mathbf w,\mathbf u)$ where $\mathbf u$ are the features we select by some feature selection algorithms. We eliminate the variables $\mathbf u$ in the distributions $Q(\mathbf w,\mathbf u|\mathbf x)$ and $P(\mathbf w,\mathbf u|\mathbf x)$. According to the equation (\ref{loss1}), the loss after elimination is $\mathcal{KL}[Q(\mathbf w|\mathbf x)\parallel P(\mathbf w|\mathbf x)]$ where $Q(\mathbf w|\mathbf x)=\int^{+\infty}_{-\infty} Q(\mathbf w,\mathbf u|\mathbf x)d\mathbf u$, $P(\mathbf w|\mathbf x)=\int^{+\infty}_{-\infty} P(\mathbf w,\mathbf u|\mathbf x)d\mathbf u$.

Subtracting the two losses and using the equation (\ref{loss1}), we get
\begin{align*}
    &\mathcal{KL}[Q(\mathbf w,\mathbf u|\mathbf x)\parallel P(\mathbf w,\mathbf u|\mathbf x)]-\mathcal{KL}[Q(\mathbf w|\mathbf x)\parallel P(\mathbf w|\mathbf x)]\\=
    &E_{Q(\mathbf w|\mathbf x)}\left(\mathcal{KL}[Q(\mathbf u|\mathbf w,\mathbf x)\parallel P(\mathbf u|\mathbf w,\mathbf x)]\right)\geq 0.
\end{align*}

The equation above demonstrates that the distributions after elimination can decrease the $\mathcal{KL}$ divergence in equation (\ref{loss1}) so that increase the ELBO. According to the inference from equation (\ref{loss1}) to equation (\ref{ELBO}), we have 

\begin{proposition} \label{marginalization}
Let $\mathbf z$ be the latent variables of VAE model and $\mathbf z=(\mathbf w,\mathbf u)$, then 
\begin{align*}
&E_{Q(\mathbf z|\mathbf x)}\left(\log P(\mathbf x|\mathbf z)\right)-\mathcal{KL}[Q(\mathbf z|\mathbf x)\parallel P(\mathbf z)]\\
\leq &E_{Q(\mathbf w|\mathbf x)}\left(\log P(\mathbf x|\mathbf w)\right)-\mathcal{KL}[Q(\mathbf w|\mathbf x)\parallel P(\mathbf w)],
\end{align*}
where 
\begin{align}\label{question}
P(\mathbf x|\mathbf w)=\int^{+\infty}_{-\infty} P(\mathbf x|\mathbf z)Q(\mathbf u|\mathbf w,\mathbf x)d\mathbf u,~ Q(\mathbf w|\mathbf x)=\int^{+\infty}_{-\infty} Q(\mathbf z|\mathbf x)d\mathbf u.
\end{align}
\end{proposition}
\begin{proof}
The detailed proof can be found in Appendix.
\end{proof}
Thus our goal of this section is to use the VAE model which is already trained to approximate the model with only $\mathbf w$ as latent variables, so that we can achieve a higher ELBO:
\begin{align}\label{ELBO elimination}
E_{Q(\mathbf w|\mathbf x)}\left(\log P(\mathbf x|\mathbf w)\right)-\mathcal{KL}[Q(\mathbf w|\mathbf x)\parallel P(\mathbf w)].
\end{align}
\textbf{Proposition \ref{marginalization}} points out a method to obtain the encoder and
decoder in the equation (\ref{ELBO elimination}). 

Fortunately, the integral operation on Gaussian distribution $Q(\mathbf z|\mathbf x)$ is simply computing the marginal distribution of $\mathbf w$. Let us assume 
\begin{align*}
    Q(\mathbf z|\mathbf x)=Q(\mathbf w,\mathbf u|\mathbf x)=\mathcal N\left(\left(\boldsymbol\mu_{\mathbf w}(\mathbf x), \boldsymbol\mu_{\mathbf u}(\mathbf x)\right),\  \begin{pmatrix}\boldsymbol\Sigma_{\mathbf w\mathbf w}(\mathbf x) & \boldsymbol\Sigma_{\mathbf w\mathbf u}(\mathbf x) \\ \boldsymbol\Sigma_{\mathbf u\mathbf w}(\mathbf x) & \boldsymbol\Sigma_{\mathbf u\mathbf u}(\mathbf x) \end{pmatrix} \right).
\end{align*}

Then the marginal distribution is
\begin{align*}
    Q(\mathbf w|\mathbf x)=\mathcal N\left(\boldsymbol\mu_\mathbf w(\mathbf x),\boldsymbol\Sigma_{\mathbf w \mathbf w}(\mathbf x)\right).
\end{align*}

To achieve this marginalization, we introduce a special $\dim\mathbf z$-dimensional  vector $\boldsymbol\pi$, whose elements are ones or zeros, to build a new model with latent variables $\mathbf z\otimes\boldsymbol\pi$.  We assume $\boldsymbol\pi=(\underset{\dim\mathbf w}{\underbrace{1,...,1}},\underset{\dim\mathbf u}{\underbrace{0,...,0}})$, and then define 
\begin{align*}
    Q(\mathbf z\otimes\boldsymbol\pi|\mathbf x)=\mathcal N\left(\left(\boldsymbol\mu_{\mathbf w}(\mathbf x), \mathbf 0_{1\times\dim\mathbf u}\right),\ \begin{pmatrix} \boldsymbol\Sigma_{\mathbf w\mathbf w}(\mathbf x) & \mathbf 0_{\dim \mathbf w\times\dim\mathbf u} \\ \mathbf 0_{\dim\mathbf u\times\dim\mathbf w} & \mathbf I_{\dim\mathbf u\times\dim\mathbf u}\end{pmatrix}\right),
\end{align*}

where $\mathbf 0_{d_1\times d_2}$ is $d_1\times d_2$ matrix whose elements are all zeros, and $\mathbf I_{d\times d}$ is $d\times d$ identity matrix.

Our aim is to derive the $\mathcal{KL}$ divergence after marginalization of encoder. So we put $Q(\mathbf z\otimes\boldsymbol\pi|\mathbf x)$ in the $\mathcal{KL}$ divergence in equation (\ref{ELBO}), then 
\begin{align*}
\mathcal{KL}[Q(\mathbf z\otimes\boldsymbol\pi|\mathbf x)\parallel P(\mathbf z)]&=\mathcal{KL}[Q(\mathbf w|\mathbf x) P(\mathbf u)\parallel P(\mathbf w)P(\mathbf u)]\\&=\mathcal{KL}[Q(\mathbf w|\mathbf x)\parallel P(\mathbf w)],
\end{align*}
which completes the marginalization of encoder in \textbf{Proposition \ref{marginalization}}. The algorithms that achieve the marginalization of $Q(\mathbf z|\mathbf x)$ with mean-field and full-covariance Gaussian posterior are shown in \textbf{Algorithm \ref{marginal mf Q}} and $\textbf{Algorithm \ref{marginal fc Q}}$

\begin{algorithm}
\renewcommand{\algorithmicrequire}{\textbf{Input:}}
\renewcommand{\algorithmicensure}{\textbf{Output:}}
\caption{The marginalization of encoder with mean-field Gaussian posterior}
\label{marginal mf Q}
\begin{algorithmic}[1]
\REQUIRE data set $\mathbb D=\{\mathbf x_1,...,\mathbf x_N\}$, VAE model, $\boldsymbol\pi$
\FOR{ data $\mathbf x$ in $\mathbb D$}
\STATE $\boldsymbol\mu$, $\log\boldsymbol\sigma^2= EncoderNetwork(\mathbf x)$;
\STATE $\boldsymbol\mu=\boldsymbol\mu\otimes\boldsymbol\pi$, $\log\boldsymbol\sigma^2=\log\boldsymbol\sigma^2\otimes\boldsymbol\pi$;
\STATE $\mathcal{KL}=\mathcal{KL}+\frac12\overset{\dim\mathbf z}{\underset{i=1}{\sum}} (\boldsymbol\mu^2+\exp(\log\boldsymbol\sigma^2)-\log\boldsymbol\sigma^2-1)_i$;
\ENDFOR
\STATE $\mathcal{KL}=\mathcal{KL}/N$;
\ENSURE $\mathcal{KL}$
\end{algorithmic}
\end{algorithm}

\begin{algorithm}
\renewcommand{\algorithmicrequire}{\textbf{Input:}}
\renewcommand{\algorithmicensure}{\textbf{Output:}}
\caption{The marginalization of encoder with full-covariance Gaussian posterior}
\label{marginal fc Q}
\begin{algorithmic}[1]
\REQUIRE data set $\mathbb D=\{\mathbf x_1,...,\mathbf x_N\}$, VAE model, $\boldsymbol\pi$
\FOR{ data $\mathbf x$ in $\mathbb D$}
\STATE $\boldsymbol\mu$, $\log\boldsymbol\sigma$, $\mathbf L_m=EncoderNetwork(\mathbf x)$;
\STATE mask matrix $\mathbf L_m$ and turn the $\mathbf L_m$ to lower triangular matrix with zeros on and above the diagonal;
\STATE $\boldsymbol\mu=\boldsymbol\mu\otimes\boldsymbol\pi$, $\log\boldsymbol\sigma=\log\boldsymbol\sigma\otimes\boldsymbol\pi$, $\mathbf L_m=\boldsymbol\pi\otimes\mathbf L_m\otimes\boldsymbol\pi^\top$;
\STATE $\mathbf L=\mathbf L_m+diag(\exp(\boldsymbol\sigma))$;
\STATE sample $\boldsymbol\epsilon$ from normal distribution $\mathcal N(0,\mathbf I)$;
\STATE $\mathbf w=\boldsymbol\mu+\boldsymbol\epsilon\mathbf L$;
\STATE $L_1=-\frac12\overset{\dim\mathbf z}{\underset{i=1}{\sum}}(\boldsymbol\epsilon^2+\log(2\pi)+\log\boldsymbol\sigma)_i$;
\STATE $L_2=-\frac12\overset{\dim\mathbf z}{\underset{i=1}{\sum}}(\mathbf w^2+\log(2\pi))_i$;
\STATE $\mathcal{KL}=\mathcal{KL}+(L_1-L_2)$;
\ENDFOR
\STATE $\mathcal{KL}=\mathcal{KL}/N$;
\ENSURE $\mathcal{KL}$
\end{algorithmic}
\end{algorithm}

The input VAE model is already trained and $\boldsymbol\pi$ in these algorithms is generated by the feature selection algorithms. The indices of zero elements in $\boldsymbol\pi$ are also the indices of elements of  $\mathbf u$ among $\mathbf z$.
These two algorithms output values of $\mathcal{KL}$ divergence after marginalization of encoder. The differences of \textbf{Algorithm \ref{marginal mf Q}} and \textbf{Algorithm \ref{marginal fc Q}} from \textbf{Algorithm \ref{mean-field}} and \textbf{Algorithm \ref{full-covariance}} are mean and variance which element-wise multiply by $\boldsymbol\pi$. In line 4 of \textbf{Algorithm \ref{marginal fc Q}}, $\boldsymbol\pi\otimes\mathbf L_m$ ($\mathbf L_m\otimes\boldsymbol\pi^\top$) represents each row (column) vector of matrix $\mathbf L_m$ element-wise multiply by $\boldsymbol\pi$ ($\boldsymbol\pi^\top$). These procedures guarantee $\mathbf u$ follow standard normal distribution and are independent with $\mathbf w$.

It should be noted that \textbf{Algorithm \ref{marginal mf Q}} and \textbf{Algorithm \ref{marginal fc Q}} seem to have little contribution on generating pictures. They are more likely giving a mathematical explanation that the ELBO we compute after optimization is the approximation of equation (\ref{ELBO elimination}).

\subsection{Weak convergence approximation of decoder}
The integral of decoder $P(\mathbf x|\mathbf z)$ in equation (\ref{question}) is much more complicated. Exact distribution after integral operations in equation (\ref{question}) is almost impossible. Therefore we need to use other ways to approximate it. Integral of decoder in equation (\ref{question}) can be transformed into expectation, i.e., 
\begin{align*}
P(\mathbf x|\mathbf w)=\int^{+\infty}_{-\infty} P(\mathbf x|\mathbf z)Q(\mathbf u|\mathbf w,\mathbf x)d\mathbf u=E_{Q(\mathbf u|\mathbf w,\mathbf x)}\left(P(\mathbf x|\mathbf z)\right).
\end{align*}
Let $\frac1K \overset{K}{\underset{j=1}{\sum}} P(\mathbf x|\mathbf w, \mathbf u_j)$ be function of random variables $\mathbf u_j$ for any given $\mathbf w$ and $\mathbf x$ where $\mathbf u_j$ is sampled i.i.d from $Q(\mathbf u|\mathbf w,\mathbf x)$.
By the strong law of large numbers, $\frac1K \overset{K}{\underset{j=1}{\sum}} P(\mathbf x|\mathbf w, \mathbf u_j)$ converges to $E_{Q(\mathbf u|\mathbf w,\mathbf x)}\left(P(\mathbf x|\mathbf z)\right)$ almost everywhere\cite{billingsley2008probability}, i.e.,
\begin{align*}
P\left(\lim_{K\to\infty} \frac1K\sum^K_{j=1}P(\mathbf x|\mathbf w,\mathbf u_j)=E_{Q(\mathbf u|\mathbf w,\mathbf x)}\left(P(\mathbf x|\mathbf z)\right)\right)=1.    
\end{align*}

Since $\mathbf u_j$ are sampled from $Q(\mathbf u|\mathbf w,\mathbf x)$ and not influenced by generated data $\mathbf x$ in decoder, then for given $\mathbf w$, we can sample a infinite sequence $\{\mathbf u_1, \mathbf u_2,...\}$ in probability 1, such that $\frac1K \overset{K}{\underset{j=1}{\sum}} P(\mathbf x|\mathbf w, \mathbf u_j)$ converges to $E_{Q(\mathbf u|\mathbf w,\mathbf x)}\left(P(\mathbf x|\mathbf z)\right)=P(\mathbf x|\mathbf w)$. Then by the measure theory\cite{billingsley2008probability}, we can definitely find a sequence $\{\mathbf u_1,\mathbf u_2,...\}$ for every $\mathbf w$ to achieve the convergence. We fix these uncountable sequences and assume projections $\{\mathbf u_1(\mathbf w),\mathbf u_2(\mathbf w),...\}$ project every $\mathbf w$ to these sequences. However if $\mathbf w$ and $\mathbf u$ are independent, these projections will become constants. Then $\frac1K \overset{K}{\underset{j=1}{\sum}} P(\mathbf x|\mathbf w, \mathbf u_j(\mathbf w))$ converges to $E_{Q(\mathbf u|\mathbf w,\mathbf x)}\left(P(\mathbf x|\mathbf z)\right)=P(\mathbf x|\mathbf w)$ point-by-point. We write $\mathbf u_j(\mathbf w)$ as $\mathbf u_j$ for short in latter easy and let $M_K(\mathbf x|\mathbf w)=\frac1K \overset{K}{\underset{j=1}{\sum}} P(\mathbf x|\mathbf w, \mathbf u_j)$ as conditional distribution of $\mathbf x$ given $\mathbf w$. 

$M_K(\mathbf x|\mathbf w)$ as a Gaussian mixture distribution will lead to a lot of computation problems if we use it directly as the approximation to $P(\mathbf x|\mathbf w)$ in ELBO. Considering $P(\mathbf x|\mathbf w)$ is a Gaussian distribution, we want to project $M_K(\mathbf x|\mathbf w)$ into the family of Gaussian distributions and use the projection as the approximation. The approximation in the same family of distributions should be much more efficient and less complex.
The following theorem proposes a way of projection using $\mathcal{KL}$ divergence.

\begin{theorem}\label{projection}
Let $M_K(\mathbf x|\mathbf w)=\frac1K \overset{K}{\underset{j}{\sum}} P(\mathbf x|\mathbf w, \mathbf u_j)=\frac1K \overset{K}{\underset{j}{\sum}} \mathcal N\left(\mathbf x|\mathbf f(\mathbf w,\mathbf u_j),\mathbf I\right)$ and $P_K(\mathbf x|\mathbf w)=\mathcal N(\boldsymbol\mu_K(\mathbf w),\mathbf I)$. Then \begin{align*}\underset{\boldsymbol\mu_K}{\arg\min} \mathcal{KL}[M_K(\mathbf x|\mathbf w)\parallel P_K(\mathbf x|\mathbf w)]=\frac1K \sum^K_{j=1} \mathbf f(\mathbf w,\mathbf u_j).
\end{align*}
We call this M-projection.
\end{theorem}
\begin{proof}
The detailed proof can be found in Appendix.
\end{proof}

Note that we use VAE to generate pictures, the value of every pixel is in interval $[0,1]$\footnote{The intervals of value of pixel are different vary from the formats of pictures, but they are all bounded. Therefore without loss of generality we can assume value of pixel lay in $[0,1]$}. Thus we usually constrain the mean of decoder $\mathbf f(\mathbf z)$ by activation function $\tanh$ or $sigmoid$ function, which means for every element in $\mathbf f(\mathbf z)$, $0\leq\big|f_i(\mathbf z)\big|\leq1$ for any $\mathbf z$. Using this assumption, we give an important lemma to explore the connection between $P(\mathbf x|\mathbf w)$ and $P_K(\mathbf x|\mathbf w)$.

\begin{Lemma}\label{exchange}
Let $P(\mathbf x|\mathbf w,\mathbf u)=\mathcal N(\mathbf f(\mathbf w,\mathbf u),\mathbf I)$. If $\mathbf f$ is bounded, i.e., there is a constant $C$ s.t. $\Vert \mathbf f(\mathbf w,\mathbf u)\Vert\leq C$ for all $\mathbf w$ and $\mathbf u$. Then for any non-negative integrable function $g(y,\mathbf x)$ with respect to $\mathbf x$ on $(-\infty,+\infty)$, and $g(y,\mathbf x)$ is monotone with respect to $y$, integral operation and limit operation are exchangeable, i.e.,
\begin{align*}
    \lim_{K\to\infty}\int^{+\infty}_{-\infty}g(M_K(\mathbf x|\mathbf w),\mathbf x)d\mathbf x=\int^{+\infty}_{-\infty}\lim_{K\to\infty}g(M_K(\mathbf x|\mathbf w),\mathbf x)d\mathbf x.
\end{align*}
\end{Lemma}
\begin{proof}
The detailed proof can be found in Appendix.
\end{proof}

\begin{figure}
\centering

\includegraphics[scale=0.6]{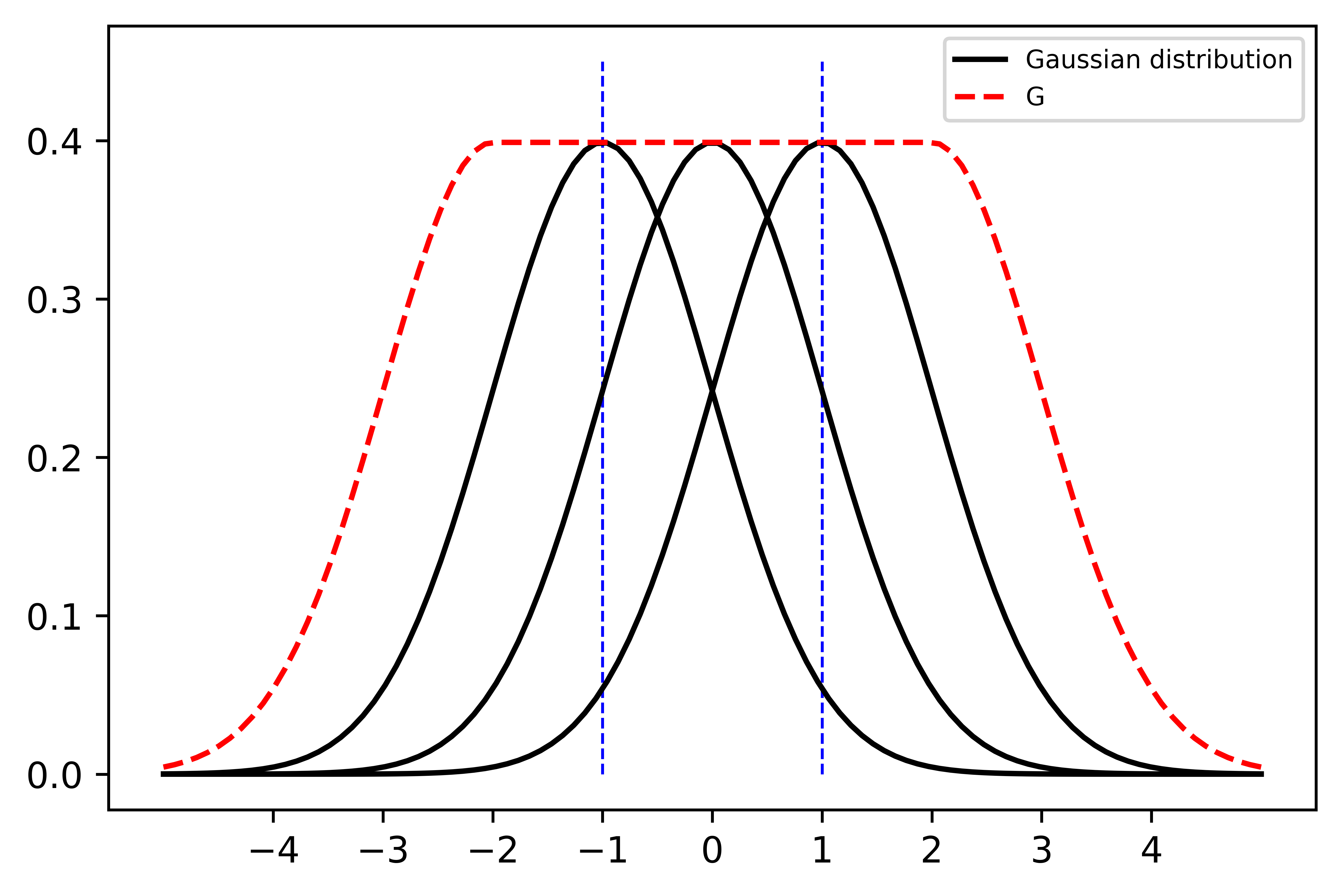}
\caption{The black lines are three examples of $P(\mathbf x|\mathbf w,\mathbf u_j)$. These three examples are Gaussian distributions with different means. Since mean function $\mathbf f(\mathbf w,\mathbf u_j)$ is bounded $\Vert\mathbf f(\mathbf w,\mathbf u_j)\Vert\leq 1$ and maximal values are all the same, we can use function $G$, the red dotted line which is also integrable on $(-\infty,+\infty)$, to dominate all $P(\mathbf x|\mathbf w,\mathbf u_j)$, then to dominate all $M_K(\mathbf x|\mathbf w)$. Then the \textbf{Lemma \ref{exchange}} is directed result of the dominated convergence theorem\cite{mcshane2013real}.}\label{dominated}
\end{figure}

The Figure \ref{dominated} states the reason why we must need to constrain the $\mathbf f(\mathbf z)$. We may use this assumption that $\mathbf f$ is bounded in this and latter section without mentioning it.
We only know $M_K(\mathbf x|\mathbf w)$ converges to $P(\mathbf x|\mathbf w)$ point-by-point, but we still need to do integral operation on $M_K(\mathbf x|\mathbf w)$. \textbf{Lemma \ref{exchange}} guarantees the exchange of limit and integral.
Now we can compute the mean and variance of $P(\mathbf x|\mathbf w)$.

\begin{theorem}\label{mean-variance}
Let random variables $\boldsymbol\xi\sim P(\mathbf x|\mathbf w)$ and $\boldsymbol\xi_K\sim M_K(\mathbf x|\mathbf w)$. Let $P_K(\mathbf x|\mathbf w)=\mathcal N(\mathbf x|\boldsymbol\mu_K(\mathbf w), \mathbf I)=\mathcal N\left(\mathbf x\bigg|\frac1K \overset{K}{\underset{j=1}{\sum}}\mathbf f(\mathbf w,\mathbf u_j),\mathbf I\right)$, then 
\begin{align*}
E(\boldsymbol\xi)=\lim_{K\to\infty}E(\boldsymbol\xi_K)=\lim_{K\to\infty}\boldsymbol\mu_K,\ \ \ \ 
D(\boldsymbol\xi)=\lim_{K\to\infty}D(\boldsymbol\xi_K)=\mathbf I.
\end{align*}
\end{theorem}
\begin{proof}
The detailed proof can be found in Appendix.
\end{proof}

By the \text{Theorem \ref{mean-variance}}, it is trivial to have that the distribution $P_K(\mathbf x|\mathbf w)$ converges to $P(\mathbf x|\mathbf w)$. But it should be noted that the convergence of mean and covariance matrix is independent with $\mathbf x$, then we can unconditionally exchange limit operation with respect to $K$ and integral on $h(P_K(\mathbf x|\mathbf w))$ related to $\mathbf x$ for any function $h$. According to the discussion, we have the theorem below.

\begin{theorem}\label{convergence in distribution}
Let random variables $\boldsymbol\eta_K \sim \mathcal N\left(\mathbf x\bigg|\frac1K \overset{K}{\underset{j=1}{\sum}}\mathbf f(\mathbf w,\mathbf u_j),\mathbf I\right)=P_K(\mathbf x|\mathbf w)$ and $\boldsymbol\xi\sim P(\mathbf x|\mathbf w)$, then $\boldsymbol\eta_K$ converges to $\boldsymbol\xi$ in distribution. Moreover, 
\begin{align*}
    \lim_{K\to\infty}\mathcal{KL}[P_K(\mathbf x|\mathbf w)\parallel P(\mathbf x|\mathbf w)]=\lim_{K\to\infty}\mathcal{KL}[P(\mathbf x|\mathbf w)\parallel P_K(\mathbf x|\mathbf w)]=0.
\end{align*}
\end{theorem}
\begin{proof}
The detailed proof can be found in Appendix.
\end{proof}

From the \textbf{Theorem \ref{convergence in distribution}}, we have a good method to approximate $P(\mathbf x|\mathbf w)$. Since By the definition of convergence in distribution, $P_K(\mathbf x|\mathbf w)$ weak converges to $P(\mathbf x|\mathbf w)$, we call this approximation as weak convergence approximation.

In the mean-field condition, latent variables $\mathbf u$ and $\mathbf w$ are independent. However in full-covariance condition, the computation of mean and covariance matrix in $Q(\mathbf u|\mathbf w,\mathbf x)$ is complex. To simplify it, we can pick $\mathbf u$ are independent with $\mathbf w$ as much as possible in feature selection process. Then $Q(\mathbf u|\mathbf w,\mathbf x)=Q(\mathbf u|\mathbf x)$ which is marginal distribution of encoder $Q(\mathbf z|\mathbf x)$. We can pick sample of $\mathbf u$ from sample of $\mathbf z$ correspond to the indices of elements of $\mathbf u$ among $\mathbf z$. Combing these above, we have the weak convergence approximation algorithms of decoder in \textbf{Algorithm \ref{approximation P mf}} and \textbf{Algorithm \ref{approximation P fc}}.

\begin{algorithm}
\renewcommand{\algorithmicrequire}{\textbf{Input:}}
\renewcommand{\algorithmicensure}{\textbf{Output:}}
\caption{Weak convergence approximation of decoder with mean-field Gaussian posterior}
\label{approximation P mf}
\begin{algorithmic}[1]
\REQUIRE data $\mathbf x$, VAE Model, $\boldsymbol\pi$, $K$
\STATE $\boldsymbol\mu$, $\log\boldsymbol\sigma^2=EncoderNetwork(\mathbf x$);
\STATE sample $\boldsymbol\epsilon$ from normal distribution $\mathcal N(0,\mathbf I)$
\STATE $\mathbf z=\boldsymbol\epsilon\otimes\exp(\frac12\log\boldsymbol\sigma^2)+\boldsymbol\mu$;
\STATE $j=0$, $\mathbf x'=0$;
\WHILE{$j<K$}
\STATE sample $\widetilde{\boldsymbol\epsilon}$ from normal distribution $\mathcal N(0,\mathbf I)$;
\STATE $\widetilde{\mathbf z}=\widetilde{\boldsymbol\epsilon}\otimes\exp(0.5\log\boldsymbol\sigma^2)+\boldsymbol\mu$;
\FOR{every index of selected feature $i$ where $\pi_i$=0}
\STATE $z_i=\widetilde{z_i}$;
\ENDFOR
\STATE $\mathbf x'=\mathbf x'+DecoderNetwork(\mathbf z)/K$;
\STATE $j=j+1$;
\ENDWHILE
\ENSURE $\mathbf x'$
\end{algorithmic}
\end{algorithm}

In \textbf{Algorithm \ref{approximation P mf}}, We input the already trained model. After getting latent variables $\mathbf z$, we generate $\widetilde{\mathbf z}$ in the same way in line 7.
By the \textbf{Theorem \ref{convergence in distribution}}, we need to sample $\mathbf u$ $K$ times from marginal distribution $Q(\mathbf u|\mathbf x)$, so we replace $z_i$ with $\widetilde{z}_i$ in line 9 where $i$ is a index of element of $\mathbf u$ among $\mathbf z$, then put the new $\mathbf z$ in the decoder network in line 11. The values of $\mathbf w$ are fixed after line 5. The $\mathbf x'$ generated after $K$ epochs is the mean of $P_K(\mathbf x|\mathbf w)$, i.e., $\mathbf x'=\frac1K\overset{K}{\underset{j=1}{\sum}} \mathbf f(\mathbf w,\mathbf u_j)$. At last, algorithm outputs the $\mathbf x'$ as the new pictures. The approximation of decoder in full-covariance condition is similar.

\begin{algorithm}
\renewcommand{\algorithmicrequire}{\textbf{Input:}}
\renewcommand{\algorithmicensure}{\textbf{Output:}}
\caption{Weak convergence approximation of decoder with full-covariance Gaussian posterior}
\label{approximation P fc}
\begin{algorithmic}[1]
\REQUIRE data $\mathbf x$, VAE Model, $\boldsymbol\pi$, $K$
\STATE $\boldsymbol\mu$, $\log\boldsymbol\sigma$, $\mathbf L_m=EncoderNetwork(\mathbf x$);
\STATE mask matrix $\mathbf L_m$ and turn the $\mathbf L_m$ to upper triangular matrix with zeros on and above the diagonal;
\STATE $\mathbf L=\mathbf L_m+diag(\exp(\boldsymbol\sigma))$;
\STATE sample $\boldsymbol\epsilon$ from normal distribution $\mathcal N(0,\mathbf I)$
\STATE $\mathbf z=\boldsymbol\mu+\boldsymbol\epsilon\mathbf L$;
\STATE $j=0$, $\mathbf x'=0$;
\WHILE{$j<K$}
\STATE sample $\widetilde{\boldsymbol\epsilon}$ from normal distribution $\mathcal N(0,\mathbf I)$;
\STATE $\widetilde{\mathbf z}=\boldsymbol\mu+\widetilde{\boldsymbol\epsilon}\mathbf L$;
\FOR{every index of selected feature $i$ where $\pi_i$=0}
\STATE $z_i=\widetilde{z_i}$;
\ENDFOR
\STATE $\mathbf x'=\mathbf x'+DecoderNetwork(\mathbf z)/K$;
\STATE $j=j+1$;
\ENDWHILE
\ENSURE $\mathbf x'$
\end{algorithmic}
\end{algorithm}

In the \textbf{Algorithm \ref{approximation P mf}} and \textbf{Algorithm \ref{approximation P fc}}, $K$ is the only hyperparameter. We need to analyse the relation between $K$ and ELBO. Firstly, let we consider $K=1$. Assumimg $\mathbf w$ and $\mathbf u$ are independent, we put $P_K(\mathbf x|\mathbf w)$ into first part of ELBO, then we have 
\begin{align*}
E_{Q(\mathbf z|\mathbf x)}\left(\log P_K(\mathbf x|\mathbf w)\right)&=E_{Q(\mathbf z|\mathbf w)}\left(\log P(\mathbf x|\mathbf w,\mathbf u_1)\right)\\
&=E_{Q(\mathbf w|\mathbf x)}\left(E_{Q(\mathbf u|\mathbf x)}\left(\log P(\mathbf x|\mathbf w,\mathbf u)\right)\right)\\
&< E_{Q(\mathbf w|\mathbf u)}\left(\log E_{Q(\mathbf u|\mathbf x)}P(\mathbf x|\mathbf w,\mathbf u)\right)\\
&=E_{Q(\mathbf w|\mathbf u)}\left(\log P(\mathbf x|\mathbf w)\right).
\end{align*}
Therefore, the condition that $K=1$ does not change the decoder. We also can find that $E_{Q(\mathbf z|\mathbf x)}\left(\log P(\mathbf x|\mathbf z)\right)$ is strictly smaller that $E_{Q(\mathbf w|\mathbf x)}\left(\log P(\mathbf x|\mathbf w)\right)$ because of strictly concave function ``$\log$".

According to the \textbf{Theorem \ref{mean-variance}} and proof of \textbf{Theorem \ref{convergence in distribution}}, we have following corollary.
\begin{corollary}\label{loss convergence}
\begin{align*}
\underset{K\to\infty}{\lim}E_{Q(\mathbf z|\mathbf x)}\left(\log P_K(\mathbf x|\mathbf w)\right)&=E_{Q(\mathbf z|\mathbf x)}\left(\underset{K\to\infty}{\lim}\log P_K(\mathbf x|\mathbf w)\right)\\&=E_{Q(\mathbf z|\mathbf x)}\left(\log P(\mathbf x|\mathbf w)\right).
\end{align*}
\end{corollary}
Thus $K>1$ is necessary, and the bigger $K$ is, the closer to equation (\ref{ELBO elimination})  we get. However every epoch needs a forward propagation, we will have too much computation if we set $K$ too big.

After all the analyses, besides requiring $\dim\mathbf w>0$ and $\dim\mathbf u>0$, we seem to have no requirement on choice of $\mathbf u$. Then why do not we just let $\dim\mathbf w=1$, $\dim\mathbf u=\dim\mathbf z-1$ and random pick one feature from $\mathbf z$ as $\mathbf w$? 
Firstly, with the increasing dimension of $\dim\mathbf u$, we need bigger $K$ to get convergence in \textbf{Corollary \ref{loss convergence}}.

\begin{theorem}\label{complexity}
Let random variable $\boldsymbol\xi\sim P(\mathbf x|\mathbf w)$.
For any $\epsilon>0$, to achieve the accuracy that $\big|E_{Q(\mathbf w|\mathbf x)}\left(\log P_K(\mathbf x|\mathbf w)\right)-E_{Q(\mathbf w|\mathbf x)}\left(\log P(\mathbf x|\mathbf w)\right)\big|<\epsilon$, weak convergence approximation algorithms need $O(K_c^{\dim\mathbf u})$ times forward propagation of encoder and decoder network, i.e., $K=O(K_c^{\dim\mathbf u})$, where $K_c$ is a constant that when $K\geq K_c$, we have
\begin{align*}
   \bigg\Vert \frac1K\sum^K_{j=1}\mathbf f(\mathbf w,\mathbf u_j)-E(\boldsymbol\xi)\bigg\Vert<\frac{\epsilon}{6},
\end{align*}
for any choice of $\mathbf u$ with $\dim\mathbf u=1$ .
\end{theorem}
\begin{proof}
The detailed proof can be found in Appendix.
\end{proof}

\textbf{Theorem \ref{complexity}} states that the complexity will increase exponentially with respect to the dimension of selected latent variables $\mathbf u$ if we want to get close enough to equation (\ref{ELBO elimination}). Secondly, not all the latent variables are worth being selected. For example, many machine learning methods, like decision trees\cite{geron2019hands}, increase the weights of important features to optimize the outcome of model. However if we increase the weights of none related features, we will get a poor outcome. Although we mentioned that we want to eliminate $\mathbf u$ in encoder and decoder, $\mathbf u$ are sampled and fed into decoder network $K$ times in \textbf{Algorithm \ref{approximation P mf}} and \textbf{Algorithm \ref{approximation P fc}} which actually increases the weight of $Q(\mathbf u|\mathbf x)$. So the more important latent variables $\mathbf u$ are, the higher the ELBO may be. And taking none related latent variables may decrease ELBO. Thus the feature selection process on latent variables $\mathbf z$ is necessary.

\section {Evidence Lower Bound Difference Score}
In this section, we propose a feature selection algorithm which is used on the latent variables of VAE specifically.  And we expand it to the classification tasks.

\begin{figure}
\centering
\includegraphics[scale=0.5]{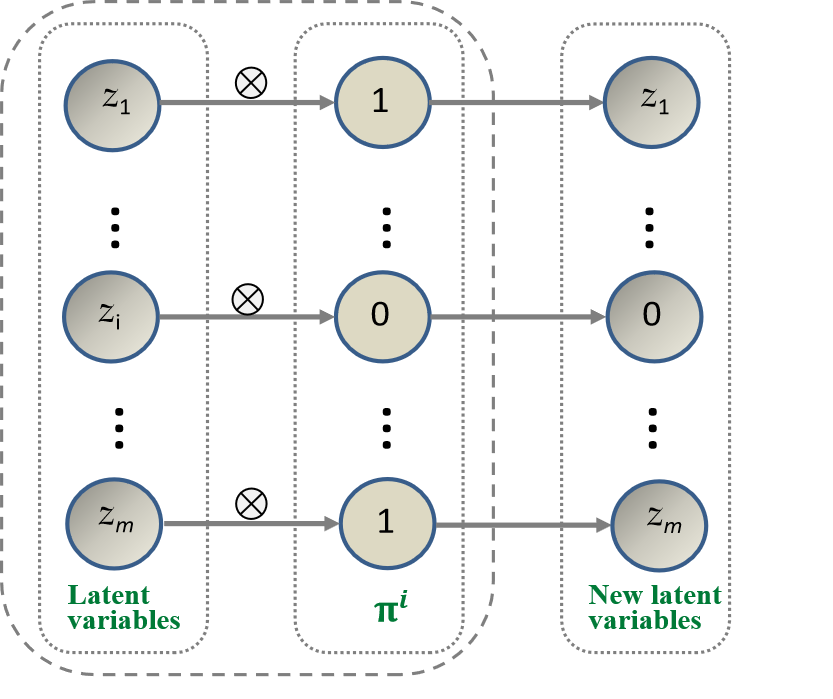}
\caption{The forward process of VAE without $i$th latent variable.}\label{pi_i}
\end{figure}

We define $\boldsymbol\pi^i$ as the $\dim\mathbf z$-dimensional row vectors whose elements are ones with zero on the $i$th element and define the ELBO without the $i$th latent variables as 
\begin{align*}
    E_{Q(\mathbf z|\mathbf x)}\left(\log P(\mathbf x|\mathbf z\otimes\boldsymbol\pi^i)\right)-\mathcal{KL}[Q(\mathbf z\otimes\boldsymbol\pi^i|\mathbf x)\parallel P(\mathbf z\otimes\boldsymbol\pi^i)],
\end{align*}
where $P(\mathbf x|\mathbf z\otimes\boldsymbol\pi^i)$ is defined as 
\begin{align*}
    P(\mathbf x|\mathbf z\otimes\boldsymbol\pi^i)=\mathcal N(\mathbf f(\mathbf z\otimes\boldsymbol\pi^i),\mathbf I).
\end{align*}
Figure \ref{pi_i} visualizes the VAE model without the $i$th latent variable. We should note that we optimize the encoder part and decoder part of ELBO separately in Section 3. It is reasonable to consider two terms in equation (\ref{ELBO}) separately. So to measure the importance of the $i$th latent variable in decoder and encoder, we subtract original ELBO and ELBO without the $i$th latent variables and add absolute symbol on each term.
\begin{align}\label{ELBD P}
    &\bigg|E_{Q(\mathbf z|\mathbf x)}\left(\log\frac{P(\mathbf x|\mathbf z)}{P(\mathbf x|\mathbf z\otimes\boldsymbol\pi^i)}\right)\bigg|\\
    \label{ELBD Q}
    +&\bigg|\mathcal{KL}[Q(\mathbf z|\mathbf x)\parallel P(\mathbf z)]-\mathcal{KL}[Q(\mathbf z\otimes\boldsymbol\pi^i|\mathbf x)\parallel P(\mathbf z\otimes\boldsymbol\pi^i)]\bigg|.
\end{align}

The equation (\ref{ELBD P}) measures the impact of the $i$th latent variable on decoder part of ELBO, and the equation (\ref{ELBD Q}) is exactly the increasing of ELBO after we keep the decoder unchanged and marginalize the encoder on the $i$th latent variable. So the bigger the score above is, the more important the $i$th latent variable is. 

The equation (\ref{ELBD Q}) can be simplified. We discuss it in different types of latent variables. In the mean-field condition, since the latent variables are independent of each other, the $\mathcal{KL}$ divergence can be separated, i.e., 
\begin{align}\label{KL mean-field}
    &\mathcal{KL}[Q(\mathbf z|\mathbf x)\parallel P(\mathbf z)]-\mathcal{KL}[Q(\mathbf z\otimes\boldsymbol\pi^i|\mathbf x)\parallel P(\mathbf z\otimes\boldsymbol\pi^i]\\ \notag
    =&\sum^{\dim\mathbf z}_{k=1} \mathcal{KL}[Q(z_k|\mathbf x)\parallel p(z_k)]-\sum^{\dim\mathbf z}_{k=1,k\neq i}\mathcal{KL}[Q(z_k|\mathbf x)\parallel P(z_k)]\\ \notag
    =&\mathcal{KL}[Q(z_i|\mathbf x)\parallel P(z_i)].
\end{align}

However in full-covariance condition, the equation (\ref{ELBD Q}) becomes quiet different. 
\begin{align}\label{KL full-covariance}
    &\mathcal{KL}[Q(\mathbf z|\mathbf x)\parallel P(\mathbf z)]-\mathcal{KL}[Q(\mathbf z\otimes\boldsymbol\pi^i|\mathbf x)\parallel P(\mathbf z\otimes\boldsymbol\pi^i)]\\ \notag
    =&E_{Q(\mathbf z_{-i}|\mathbf x)}\left(\mathcal{KL}[Q(z_i|\mathbf z_{-i},\mathbf x)\parallel P(z_i)]\right),
\end{align}
where $\mathbf z_{-i}$ is $\mathbf z$ without the $i$th element and $\dim\mathbf z_{-i}=\dim\mathbf z-1$. According to the expression of conditional partitioned Gaussian distribution\cite{bishop2006pattern}, we compute the $\mathcal{KL}[Q(z_i|\mathbf z_{-i},\mathbf x)\parallel P(z_i)]$ using following theorem. 
\begin{theorem}\label{CPGD}
Let $(\mathbf w,\mathbf u)\sim\mathcal N(0,\mathbf I)$, $(\mathbf w,\mathbf u)|\mathbf x\sim \mathcal N(\boldsymbol\mu,\boldsymbol\Sigma)$ where $\boldsymbol\mu=(\boldsymbol\mu_\mathbf w,\boldsymbol\mu_\mathbf u)$. And let 
\begin{align*}
    \boldsymbol\Lambda=\boldsymbol\Sigma^{-1}=\begin{pmatrix} \boldsymbol\Lambda_{\mathbf w\mathbf w} & \boldsymbol\Lambda_{\mathbf w\mathbf u} \\ \boldsymbol\Lambda_{\mathbf u\mathbf w} & \boldsymbol\Lambda_{\mathbf u\mathbf u}\end{pmatrix}.
\end{align*}
Let $\boldsymbol\Sigma_{\mathbf u|\mathbf w}=\boldsymbol\Lambda_{\mathbf u\mathbf u}^{-1}$ and $\boldsymbol\mu_{\mathbf u|\mathbf w}=\boldsymbol\mu_\mathbf u-(\mathbf w-\boldsymbol\mu_\mathbf w)\boldsymbol\Lambda_{\mathbf w\mathbf u}\boldsymbol\Sigma_{\mathbf u|\mathbf w}$, then
\begin{align*}
    \mathcal{KL}[Q(\mathbf u|\mathbf w,\mathbf x)\parallel P(\mathbf u)]&=\frac12\sum^{\dim\mathbf u}_{i=1}\left((\boldsymbol\mu_{\mathbf u|\mathbf w})_i+(\boldsymbol\Sigma_{\mathbf u|\mathbf w})_{ii}-1\right)\\
    &-\frac12\log\det\big|\boldsymbol\Sigma_{\mathbf w\mathbf w}\big|,
\end{align*}
where $(\boldsymbol\Sigma_{\mathbf u|\mathbf w})_{ii}$ is the $i$th element in diagonal of $\boldsymbol\Sigma_{\mathbf u|\mathbf w}$.
\end{theorem}
\begin{proof}
The detailed proof can be found in Appendix.
\end{proof}

If we want to compute $\mathcal{KL}[Q(z_i|\mathbf z_{-i},\mathbf x)\parallel P(z_i)]$, we only need to let $(\mathbf w,\mathbf u)=(\mathbf z_{-i},z_i)$ and $(\boldsymbol\mu,\boldsymbol\Sigma)=(\boldsymbol\mu(\mathbf x),\boldsymbol\Sigma(\mathbf x))$ in \textbf{Theorem \ref{CPGD}}.

Recall that in the weak convergence approximation algorithms, we need $\mathbf w$ to be independent of $\mathbf u$ as much as possible. Thus according to the definition of mutual information\cite{li2017feature}, we define the mutual information of the $i$th latent variable as $\mathcal{KL}[Q(\mathbf z_{-i},z_i|\mathbf x)\parallel Q(\mathbf z_{-i}|\mathbf x)Q(z_i|\mathbf x)]$. This term measures the amount of information shared by $\mathbf z_{-i}$ and $z_i$, and is equal to zero if and only if $\mathbf z_{-i}$ and $z_i$ are independent. So we minus mutual information as a penalty term in full-covariance condition. Furthermore, we have the following theorem,

\begin{theorem} Let mutual information $MI=\mathcal{KL}[Q(\mathbf z|\mathbf x)\parallel Q(\mathbf w|\mathbf x)Q(\mathbf u|\mathbf x)]$, then 
\begin{align*}0\leq E_{Q(\mathbf w|\mathbf x)}\left(\mathcal{KL}[Q(\mathbf u|\mathbf w,\mathbf x)\parallel P(\mathbf u)]\right)-MI\leq \mathcal{KL}[Q(\mathbf z|\mathbf x)\parallel P(\mathbf z)].
\end{align*}
And maximizing the $E_{Q(\mathbf w|\mathbf x)}\left(\mathcal{KL}[Q(\mathbf u|\mathbf w,\mathbf x)\parallel P(\mathbf u)]\right)-MI$ with respect to different choices of $\mathbf u$ is equivalent to minimize $ E_{Q(\mathbf u|\mathbf x)}\left(\mathcal{KL}[Q(\mathbf w|\mathbf u,\mathbf x)\parallel P(\mathbf w)]\right)$.
\end{theorem}
\begin{proof}
The detailed proof can be found in Appendix.
\end{proof}

Since $E_{Q(\mathbf u|\mathbf x)}\mathcal{KL}\left([Q(\mathbf w|\mathbf u,\mathbf x)\parallel P(\mathbf w)]\right)$ is zero if and only if $Q(\mathbf w|\mathbf u,\mathbf x)$ is equal to standard normal distribution $P(\mathbf w)$ for any $\mathbf u$. Thus maximizing $E_{Q(\mathbf w|\mathbf x)}\left(\mathcal{KL}[Q(\mathbf u|\mathbf w,\mathbf x)\parallel P(\mathbf u)]\right)-MI$ not only can guarantee the independence of $\mathbf w$ and $\mathbf u$, but also states that $\mathbf u$ contain more information and are more important than $\mathbf w$. 

Summarizing all the discussion above, we define the evidence lower bound difference (ELBD) score of the $i$th latent variable as
\begin{align} \label{final ELBD}
\left\{
\begin{aligned}
&\text{mean-filed}: \bigg|E_{Q(\mathbf z|\mathbf x)}\left(\log\frac{P(\mathbf x|\mathbf z)}{P(\mathbf x|\mathbf z\otimes\boldsymbol\pi^i)}\right)\bigg|+\mathcal{KL}[Q(z_i|\mathbf x)\parallel P(z_i)]; \\
&\text{full-covariance}: \bigg|E_{Q(\mathbf z|\mathbf x)}\left(\log\frac{P(\mathbf x|\mathbf z)}{P(\mathbf x|\mathbf z\otimes\boldsymbol\pi^i)}\right)\bigg|\\
&\qquad\qquad\qquad+E_{Q(\mathbf z_{-i}|\mathbf x)}\left(\mathcal{KL}[Q(z_i|\mathbf z_{-i},,\mathbf x)\parallel P(z_i)]\right)\\
&\qquad\qquad\qquad-\mathcal{KL}[Q(\mathbf z_{-i},z_i|\mathbf x)\parallel Q(\mathbf z_{-i}|\mathbf x)Q(z_i|\mathbf x)].\\
\end{aligned}
\right.
\end{align}
Actually, ELBD score in mean-field condition is a special case of ELBD score in full-covariance condition. As all the latent variables are independent in mean-field condition, the mutual information of every latent variable is always zero.
Finally, we give the algorithms to compute these two kinds of ELBD scores.

\begin{algorithm}
\renewcommand{\algorithmicrequire}{\textbf{Input:}}
\renewcommand{\algorithmicensure}{\textbf{Output:}}
\caption{Computing all the ELBD scores in mean-field Gaussian posterior condition}
\label{ELBD mean-field}
\begin{algorithmic}[1]
\REQUIRE data set $\mathbb D=\{\mathbf x_1,...,\mathbf x_N\}$, VAE model
\STATE ELBD=\{\};
\FOR{$i$ from 1 to $\dim\mathbf z$}
\STATE $L=0$, $\mathcal{KL}=0$;
\STATE Let $\boldsymbol\pi^i=(1,1,...,0,...,1)$ where $0$ is the $i$th element; 
\FOR{ data $\mathbf x$ in $\mathbb D$}
\STATE $\boldsymbol\mu$, $\log\boldsymbol\sigma^2=EncoderNetwork(\mathbf x$);
\STATE sample $\boldsymbol\epsilon$ from normal distribution $\mathcal N(0,\mathbf I)$
\STATE $\mathbf z=\boldsymbol\epsilon\otimes\exp(\frac12\log\boldsymbol\sigma^2)+\boldsymbol\mu$;
\STATE $\mathbf x'=DecoderNetwork(\mathbf z)$;
\STATE $\widetilde{\mathbf x}=DecoderNetwork(\mathbf z\otimes\boldsymbol\pi^i)$;
\STATE $L=L+\frac12\underset{k=1}{\overset{\dim\mathbf x}{\sum}}\left(\big|(\mathbf x-\mathbf x')^2-(\mathbf x-\widetilde{\mathbf x})^2\big|\right)_k$;
\STATE $kl=\frac12\left(\mu^2_i+\exp(\log\sigma^2_i)-\log\sigma^2_i-1\right)$
\STATE $\mathcal{KL}=\mathcal{KL}+kl$
\ENDFOR
\STATE $ELBD_i=(L+\mathcal{KL})/N$;
\STATE add $ELBD_i$ in the set ELBD;
\ENDFOR
\ENSURE ELBD
\end{algorithmic}
\end{algorithm}

In \textbf{Algorithm \ref{ELBD mean-field}} and \textbf{Algorithm \ref{ELBD full-covariance}}, we can only choose a mini-batch of data set as input data set $\mathbb D$ to reduce computation. In \textbf{Algorithm \ref{ELBD mean-field}}, line 11 computes equation (\ref{ELBD P}) of the $i$th latent variable for every pair of original data $\mathbf x$ and generated data $\mathbf x'$, $\widetilde{\mathbf x}$. The $L$ is the unbiased estimation of equation (\ref{ELBD P}). Line 12 computes the exact value of equation (\ref{KL mean-field}). Adding these two values and divide the number of data, we obtain the unbiased estimation of ELBD in mean-field condition in equation (\ref{final ELBD}) with respect to $Q(\mathbf z|\mathbf x)$.

In \textbf{Algorithm \ref{ELBD full-covariance}}, we also compute the exact value $\mathcal{KL}$ divergence in line 14. However it is the unbiased estimation of equation (\ref{KL full-covariance}). Meanwhile $L_1-L_2-L_3$ is the unbiased estimation of mutual information where $\boldsymbol\Sigma_{-i}$ is covariance matrix that removes the $i$th row and the $i$th column. After all we have the unbiased estimation of ELBD score of the $i$th latent variable under full-covariance condition  in line 22.

\begin{algorithm}
\renewcommand{\algorithmicrequire}{\textbf{Input:}}
\renewcommand{\algorithmicensure}{\textbf{Output:}}
\caption{Computing all the ELBD scores in full-covariance Gaussian posterior condition}
\label{ELBD full-covariance}
\begin{algorithmic}[1]
\REQUIRE data set $\mathbb D=\{\mathbf x_1,...,\mathbf x_N\}$, VAE model
\STATE ELBD=\{\};
\FOR{$i$ from 1 to $\dim\mathbf z$}
\STATE $L=0$, $\mathcal{KL}=0$, $MI=0$;
\STATE Let $\boldsymbol\pi^i=(1,1,...,0,...,1)$ where $0$ is the $i$th element; 
\FOR{ data $\mathbf x$ in $\mathbb D$}
\STATE $\boldsymbol\mu$, $\log\boldsymbol\sigma$, $\mathbf L_m=EncoderNetwork(\mathbf x$);
\STATE mask matrix $\mathbf L_m$ and turn the $\mathbf L_m$ to lower triangular matrix with zeros on and above the diagonal;
\STATE $\mathbf L=\mathbf L_m+diag(\exp(\boldsymbol\sigma))$;
\STATE sample $\boldsymbol\epsilon$ from normal distribution $\mathcal N(0,\mathbf I)$
\STATE $\mathbf z=\boldsymbol\mu+\boldsymbol\epsilon\mathbf L$;
\STATE $\mathbf x'=DecoderNetwork(\mathbf z)$;
\STATE $\widetilde{\mathbf x}=DecoderNetwork(\mathbf z\otimes\boldsymbol\pi^i)$;
\STATE $L=L+\frac12\underset{k=1}{\overset{\dim\mathbf x}{\sum}}\left(\big|(\mathbf x-\mathbf x')^2-(\mathbf x-\widetilde{\mathbf x})^2\big|\right)_k$;
\STATE Compute $kl$ using \textbf{Theorem \ref{CPGD}};
\STATE $\mathcal{KL}=\mathcal{KL}+kl$;
\STATE $\boldsymbol\Sigma=\mathbf L\mathbf L^\top$;
\STATE $L_1=-\frac12(\mathbf z-\boldsymbol\mu)\boldsymbol\Sigma^{-1}(\mathbf z-\boldsymbol\mu)^\top$;
\STATE $L_2=-\frac12(\mathbf z_{-i}-\boldsymbol\mu_{-i})\boldsymbol\Sigma_{-i}^{-1}(\mathbf z_{-i}-\boldsymbol\mu_{-i})^\top$;
\STATE $L_3=-\frac12(z_i-\mu_i)(\boldsymbol\Sigma)_{ii}^{-1}(z_i-\mu_i)$;
\STATE $MI=MI+L_1-L_2-L_3$;
\ENDFOR
\STATE $ELBD_i=(L+\mathcal{KL}-MI)/N$;
\STATE add $ELBD_i$ in the set ELBD;
\ENDFOR
\ENSURE ELBD
\end{algorithmic}
\end{algorithm}

After having ELBD scores for all latent variables, we can decide how many features we need to select begin at the largest ELBD score to the smallest. Then we build zero-one vector $\boldsymbol\pi$ by setting the values as zeros correspond to the indices of selected features $\mathbf u$ and others as ones. 

Besides ELBD, we can use many other feature selection algorithms to build $\boldsymbol\pi$. For each data $\mathbf x$, we use encoder network and reparameterization to get $\mathbf z$, and see $\mathbf z$ as a new data. So we turn the data set $\{\mathbf x_1,...,\mathbf x_N\}$ to $\{\mathbf z_1,...,\mathbf z_N\}$, and do the feature selection on the new data set. Similarly, we pick the important features and set elements of $\boldsymbol\pi$ as zeros with respect to the indices of important features.

We can extend the ELBD score such that ELBD score can not only measure the importance of latent variables in VAE, but can select features on many classification tasks. Now let consider classification task data set $\mathbb D_c=\{(\mathbf x_1,\mathbf y_1),...,(\mathbf x_N,\mathbf y_N)\}$. Let $M$ be the model which is already trained on $\mathbb D_c$ without feature selection process and $M(\mathbf x)$ is the prediction $M$ makes on data $\mathbf x$. The type of $M$ should be the same as the model we want to use on $\mathbb D_c$ after feature selection process. For example, if we want to use the decision tree model after feature selection, we will choose $M$ as decision tree model and train it on $\mathbb D_c$ without feature selection process.  We define the generalized ELBD (gELBD) score of the $i$th feature as 
\begin{align}\label{gELBD}
    \frac1N\sum^{N}_{k=1}\left(\big |L(\mathbf y_k,\mathbf y_k')-L(\mathbf y_k,\mathbf y_k^i)\big|\right)
    +\frac1N\sum^{N}_{k=1}\sum_{x_j\in Val(i)} P(x_j|\mathbf y_k)\log\frac{P(x_j|\mathbf y_k)}{P(x_j)},
\end{align}
where $L$ is the loss function which is used during the training process of $M$ and $\mathbf y_k'=M(\mathbf x_k)$, $\mathbf y_k^i=M(\mathbf x_k\otimes\boldsymbol\pi^i)$. $Val(i)$ is the range that the $i$th feature can take. The probability $P(x_j)$ stands for the frequency of value $x_j$ in $\mathbb D_c$ and $P(x_j|\mathbf y_k)$ represents the frequency of value $x_j$ with label $\mathbf y_k$. Therefore the gELBD can be only used on discrete features with discrete labels and if we want use it on continuous features, some data discretization techniques will be needed beforehand. 

Compared to ELBD, gELBD (\ref{gELBD}) is nearly the copy of equation (\ref{ELBD P}) and (\ref{ELBD Q}), and it is same as mean-field condition in equation (\ref{final ELBD}). However gELBD sees $\mathbf z$ as original data and $\mathbf x$ as label of $\mathbf z$, and then $\mathbf x'$ becomes the prediction of model $M$. And we do not make assumption that every feature is independent of each other. The second term of equation (\ref{gELBD}) has the equivalent form that
\begin{align*}
    &\frac1N\sum^{N}_{k=1}\sum_{x_j\in Val(i)} P(x_j|\mathbf y_k)\log\frac{P(x_j|\mathbf y_k)}{P(x_j)}\\
    =&\sum_{\mathbf y_k}P(\mathbf y_k)\sum_{x_j\in Val(i)} P(x_j|\mathbf y_k)\log\frac{P(x_j|\mathbf y_k)}{P(x_j)},
\end{align*}
which is also called information gain in other articles\cite{li2017feature}.

Different from other feature selection scores, ELBD and gELBD need to train models without feature selection process beforehand, and they need the prediction on $\mathbf x\otimes\boldsymbol\pi^i$ for all data $\mathbf x$ and $1\leq i\leq\dim\mathbf z$. Thus ELBD and gELBD are more complex that they need $O(\dim\mathbf z\times N)$ times forward computation of model. However we may see in Section 5 that ELBD and gELBD have better performance than many other features selection algorithms including state-of-the-art feature selection algorithms.

\section{Experimental Studies and Discussions}
In this section, we use 7 data sets to evaluate our algorithms, including MNIST\cite{MNIST}, KMNIST\cite{clanuwat2018deep}, Hands Gestures\cite{HandGestures}, Fingers\cite{Fingers}, Brain\cite{Brain}, Yale\cite{Yale} and Chest\cite{Chest}. The basic information about these data sets is listed in Table \ref{data information}. We also provide the values of some hyperparameters in Table \ref{data information}.  We use 4 types of VAE: VAE, CVAE, NF-VAE, IAF-VAE, which are mentioned in Section 2, to fit these 7 data sets. Then we choose $\mathbf u$ from the latent variables and use algorithms in Section 3 to optimize models under different types of Gaussian posterior, i.e., mean-field and full-covariance. To select important latent variables as $\mathbf u$, we apply ELBD algorithm. We also use other 9 feature selection algorithms, Laplacian score (Lap)\cite{hastie2009elements}, spectral feature selection (SPEC)\cite{zhao2007spectral},  multi-cluster feature selection (MCFS)\cite{cai2010unsupervised}, nonnegative discriminative feature selection (NDFS)\cite{li2012unsupervised}, unsupervised feature selection (UDFS)\cite{yang2011l2}, infinite feature selection (Inf)\cite{9119168}, fisher score (Fisher)\cite{hart2000pattern}, efficient and robust feature selection (RFS)\cite{nie2010efficient} and relevant feature selection (ReliefF)\cite{robnik2003theoretical}, to select $\mathbf u$ and to compare with ELBD. These 9 algorithms are commonly used and Inf is one of the state-of-the-art feature selection algorithms. Furthermore we get other 5 data sets, Fashion-MNIST\cite{xiao2017fashion}, EMNIST\cite{cohen2017emnist}, Shape\cite{Shapes}, Chinese Calligraphy\cite{CC} and Eyes\cite{Eyes}, which are all picture data sets. The basic information about these data sets is also shown in Table \ref{classification information}. Then we use gELBD to do the feature selection on classification task on these data sets and we also use these 9 feature selection algorithms as comparisons. We do all these experiments on GPUs 3060 and 2080Ti with 16G memory. The code of algorithms is available in \href{https://github.com/ronedong/ELBD-Efficient-score-algorithm-for-feature-selection-on-latent-variables-of-VAE}{https://github.com/ronedong/ELBD-Efficient-score-algorithm-for-feature-selection-on-latent-variables-of-VAE}

\subsection{Optimize results of VAE}

\begin{table*}[htbp]
\centering
\fontsize{9}{13}\selectfont
\caption{Basic information about data sets of generative tasks}\label{data information}
\setlength{\tabcolsep}{3mm}{
\begin{tabular}{|c|c|c|c|c|c|c|c|}
\hline
\multirow{2}{*}{Datasets} & Number of  & Number of pixels &  \multirow{2}{*}{Categories}     & Train set size vs & Dimension of         & Mini-batch &\multirow{2}{*}{$K$}\cr   
                          & pictures   & on one picture   &                                  & Test set size   
            & latent variables        & size       & 
\cr\hline 
\hline
MNIST          & 70000  & 28$\times$28     & 10  & 6:1  & 50   & 2000 & 15 \cr\hline
KMNIST         & 70000  & 28$\times$28     & 10  & 6:1  & 50   & 2000 & 15 \cr\hline
Hand Gestures  & 24000  & 50$\times$50     & 20  & 3:1  & 100  & 1000 & 10 \cr\hline
Fingers        & 21600  & 64$\times$64     & -   & 5:1  & 100  & 600  & 10 \cr\hline
Brain          & 4600   & 128$\times$128   & 2   & 4:1  & 150  & 300  & 10  \cr\hline
Yale           & 2452   & 150$\times$150   & -   & 4:1  & 200  & 200  & 10  \cr\hline
Chest          & 5856   & 150$\times$150   & 2   & 8:1  & 150  & 200  & 10  \cr\hline
\end{tabular}}
\end{table*}

\begin{table*}[htbp]
\centering
\fontsize{6}{13.5}\selectfont
\caption{MNIST}\label{MNIST}
\setlength{\tabcolsep}{2mm}{
\begin{tabular}{|c|c|c|c|c|c|c|c|c|c|c|c|c|}
\hline

Model & Loss & origin & ELBD & Lap & SPEC & MCFS & NDFS & UDFS & Inf & Fisher & RFS & ReliefF \cr\hline
\hline
\multirow{2}{*}{VAE(MF)}     & -ELBO & 69.3(0.93) & \textbf{31.5(0.53)} & 32.5(0.38) & 32.6(0.43) & 57.4(2.72) & 64.2(0.80) & 32.0(1.07) & 31.6(0.45) & 31.8(0.43) & 38.2(1.51) & 31.8(0.47) 
\cr\cline{2-13}
                             & L2   & 40.5(1.53) & \textbf{31.3(0.45)} & 33.3(0.44) & 32.3(0.47) & 37.2(1.22) & 38.1(1.15) & 31.5(0.47) & 32.9(0.43) & 31.7(0.40) & 37.9(0.86) & 31.7(0.54) 
\cr\hline
\multirow{2}{*}{VAE(FC)}     & -ELBO & 29.0(0.12) & \textbf{5.7(0.09)}  & 20.6(0.17) & 16.8(0.55) & 9.0(0.85)  & 6.5(0.29)  & 12.5(0.90) & 11.6(1.24) & 5.9(0.16)  & 6.2(0.08)  & 8.9(0.83) \cr\cline{2-13}
                             & L2   & 35.6(0.45) & \textbf{26.5(0.13)} & 34.9(0.30) & 34.4(0.15) & 29.6(0.73) & 27.2(0.18) & 32.4(0.50) & 31.4(0.96) & 26.6(0.16) & 27.0(0.10) & 30.2(0.79) 
\cr\hline
\multirow{2}{*}{CVAE(MF)}    & -ELBO & 67.0(1.54) & \textbf{42.4(1.47)} & 43.4(1.42) & 43.4(1.50) & 59.2(4.15) & 65.9(0.92) & 43.5(1.55) & 43.4(1.64) & 43.5(1.55) & 66.2(1.23) & 43.6(1.42 \cr\cline{2-13}
                             & L2   & 48.5(1.57) & \textbf{42.4(1.47)} & 43.3(1.42) & 43.3(1.47) & 46.1(1.44) & 47.9(1.56) & 43.3(1.46) & 43.3(1.49) & 43.3(1.39) & 47.9(1.61) & 43.4(1.44) 
\cr\hline
\multirow{2}{*}{CVAE(FC)}    & -ELBO & 29.2(0.54) & \textbf{4.2(0.09)}  & 33.3(1.4) & 32.4(1.73) & 29.5(3.21) & 29.4(2.3) & 28.8(0.73) & 29.0(1.29) & 19.9(0.59) & 25.9(1.38) & 36.9(1.86)   \cr\cline{2-13}
                             & L2   & 35.2(0.05) & \textbf{25.8(0.05)} & 80.9(2.5) & 79.1(2.9) & 74.8(6.42) & 74.4(4.63) & 72.9(1.22) & 73.9(2.41) & 56.6(0.94) & 67.9(2.12) & 86.3(3.88)   
\cr\hline
\multirow{2}{*}{NF-VAE(MF)}  & -ELBO & 71.8(1.25) & \textbf{32.4(1.06)} & 32.5(1.11) & \textbf{32.4(1.23)} & 57.0(1.33) & 64.9(0.88) & 32.6(1.05) & 33.0(1.24) & \textbf{32.4(0.97)} & 61.0(1.71) & 32.8(1.09) 
\cr\cline{2-13}
                             & L2   & 41.4(1.42) & \textbf{31.6(0.14)} & 32.4(1.17) & 32.3(1.17) & 37.1(1.6) & 38.5(1.32) & 32.4(1.14) & 32.5(1.17) & 32.2(1.08) & 38.4(1.28) & 32.5(1.14)  
\cr\hline
\multirow{2}{*}{NF-VAE(FC)}  & -ELBO & 29.1(0.02) & \textbf{6.3(0.23)}  & 20.7(0.2) & 23.3(1.05) & 12.0(1.22) & 7.6(0.68) & 19.1(2.74) & 16.5(1.04) & 6.5(0.09) & 7.6(0.05) & 10.2(0.17)  
\cr\cline{2-13}
                             & L2   & 36.2(0.43) & \textbf{27.5(0.31)} & 35.5(0.55) & 47.5(2.18) & 35.9(2.53) & 29.0(0.82) & 45.0(4.18) & 39.3(2.2) & 27.7(0.09) & 29.1(0.01) & 32.5(0.2)  
\cr\hline
\multirow{2}{*}{IAF-VAE(MF)} & -ELBO & 18.44(0.24) & \textbf{18.39(0.21)} & \textbf{18.39(0.22)} & 18.41(0.22) & 18.41(0.21) & 18.41(0.22) & 18.40(0.22) & 18.42(0.22) & 18.42(0.22) & 18.40(0.22) & 18.44(0.24)
\cr\cline{2-13}
                             & L2   & 18.42(0.22) & \textbf{18.39(0.22)} & \textbf{18.39(0.22)} & 18.40(0.22) & 18.40(0.21) & 18.41(0.23) & \textbf{18.39(0.21)} & 18.41(0.23) & 18.40(0.21) & 18.40(0.22) & 18.42(0.22) 
\cr\hline

\multirow{2}{*}{IAF-VAE(FC)} & -ELBO & 14.4(1.03) & \textbf{7.9(0.96)} & 8.8(0.73) & 8.8(0.77) & 8.3(1.0) & 8.4(0.77) & 8.7(1.2) & 8.2(0.93) & 8.0(0.93) & 8.0(0.95) & 8.3(0.9)  
\cr\cline{2-13}
                             & L2   & 34.2(1.73) & \textbf{33.4(1.91)} & 34.0(1.71) & 34.0(1.72) & 33.7(1.96) & 33.7(1.75) & 33.9(1.95) & 33.6(1.87) & \textbf{33.4(1.89)} & \textbf{33.4(1.89)} & 33.6(1.85)
\cr\hline
\end{tabular}}
\end{table*}

\begin{table*}[htbp]
\centering
\fontsize{6}{13.5}\selectfont
\caption{KMNIST}\label{KMNIST}
\setlength{\tabcolsep}{2mm}{
\begin{tabular}{|c|c|c|c|c|c|c|c|c|c|c|c|c|}
\hline

Model & Loss & origin & ELBD & Lap & SPEC & MCFS & NDFS & UDFS & Inf & Fisher & RFS & ReliefF \cr\hline
\hline
\multirow{2}{*}{VAE(MF)}     & -ELBO & 149.4(1.59) & \textbf{99.8(2.05)} & 110.2(1.16) & 107.9(1.36) & 106.5(4.68) & 115.5(1.65) & 105.3(1.25) & 101.7(2.09) & 100.5(1.68) & 104.4(3.53) & \textbf{99.8(1.89)}
\cr\cline{2-13}
                             & L2   & 99.7(1.69) & \textbf{87.5(1.57)}  & 88.6(1.67) & 88.9(2.09) & 88.6(2.28) & 89.0(1.83) & 88.44(2.01) & 88.4(1.75) & 88.0(1.9) & 88.9(2.15) & \textbf{87.5(2.07)} 
\cr\hline
\multirow{2}{*}{VAE(FC)}     & -ELBO & 67.2(0.09) & 41.2(0.17) & 46.9(0.14) & 43.5(0.42) & 41.5(0.22) & 41.2(0.35) & 43.8(0.34) & 46.3(0.41) & \textbf{40.5(0.11)} & 41.0(0.15) & 43.6(0.5) \cr\cline{2-13}
                             & L2   & 93.6(0.25) & 84.7(0.29) & 84.7(0.52) & 84.5(0.6) & 83.5(0.63) & 83.7(0.48) & 84.4(0.8) & 84.7(0.58) & \textbf{83.4(0.68)} & 83.9(0.51) & 84.8(0.6) 
\cr\hline
\multirow{2}{*}{CVAE(MF)}    & -ELBO & 161.8(0.65) & \textbf{127.0(0.6)} & 127.3(0.64) & 127.3(0.6) & 150.0(3.35) & 159.8(1.03) & 127.3(0.58) & 127.5(0.64) & 127.8(1.23) & 160.9(1.08) & 127.9(0.62) 
\cr\cline{2-13}
                             & L2   & 136.7(1.77) & \textbf{126.9(0.56)}  & 127.3(0.68) & 127.2(0.55) & 132.6(1.72) & 135.3(1.16) & 127.3(0.66) & 127.4(0.65) & 127.4(0.96) & 135.5(1.54) & \textbf{126.9(0.68)}  
\cr\hline
\multirow{2}{*}{CVAE(FC)}    & -ELBO & 84.1(0.15) & \textbf{68.3(0.16)} & 90.3(0.86) & 91.3(1.07) & 83.2(0.15) & 80.3(2.79) & 86.9(1.89) & 82.6(1.68) & 82.0(2.04) & 87.6(1.33) & 84.9(1.4) \cr\cline{2-13}
                             & L2   & 159.9(0.39) & \textbf{156.8(0.31)} & 193.6(1.5) & 195.8(1.51) & 181.3(0.36) & 176.2(5.22) & 188.2(3.96) & 181.0(3.35) & 179.3(3.53) & 190.5(2.52) &  182.5(1.98)   
\cr\hline
\multirow{2}{*}{NF-VAE(MF)}  & -ELBO & 152.7(0.82) & \textbf{103.1(0.9)} & 114.6(1.1) & 112.0(0.73) & 109.1(1.81) & 117.1(0.45) & 106.8(1.15) & 105.9(0.71) & 104.7(0.73) & 107.1(0.75) &  103.2(0.46)
\cr\cline{2-13}
                             & L2   & 103.6(0.8) & \textbf{90.6(0.86)}  & 92.2(1.0) & 92.1(1.39) & 90.8(1.04) & 92.0(1.35) & 90.7(0.27) & 90.8(0.85) & 91.6(0.9) & 91.2(0.64) & 99.7(0.39) 
\cr\hline
\multirow{2}{*}{NF-VAE(FC)}  & -ELBO & 67.2(0.09) & \textbf{41.2(0.2)}   & 46.8(0.15) & 47.3(1.68) & 43.3(0.84) & 42.2(0.43) & 46.5(1.56) & 46.4(0.15) & 42.9(0.21) & 43.4(0.18) & 45.1(0.5)  
\cr\cline{2-13}
                             & L2   & 93.7(0.29) & \textbf{84.7(0.29)}  & 85.2(0.38) & 92.5(2.65) & 87.5(1.32) & 85.6(0.47) & 90.6(3.78) & 85.2(0.35) & 86.2(0.49) & 86.8(0.39) & 88.4(1.02)  
\cr\hline
\multirow{2}{*}{IAF-VAE(MF)} & -ELBO & 88.7(0.77) & \textbf{84.8(0.03)} & 85.8(0.78) & 85.7(0.84) & 88.0(1.03) & 88.7(0.97) & 85.8(0.77) & 85.9(0.76) & 85.8(0.93) & 89.0(1.04) & 85.4(0.85) 
\cr\cline{2-13}
                             & L2   & 81.4(0.93) & \textbf{77.9(0.88)} & 78.3(0.85) & 78.3(0.95) & 80.6(1.14) & 81.4(1.03) & 78.3(0.83) & 78.4(0.8) & 78.3(0.92) & 81.3(1.01) & \textbf{77.9(0.85)} 
\cr\hline

\multirow{2}{*}{IAF-VAE(FC)} & -ELBO & 61.2(1.35) & \textbf{50.7(1.64)} & 55.4(1.71) & 55.4(1.66) & 52.6(0.99) & 51.4(1.43) & 53.6(2.01) & 51.0(1.64) & 51.4(0.93) & 52.5(1.45) & 50.9(1.61)
\cr\cline{2-13}
                             & L2   & 121.4(3.16) & \textbf{118.2(3.28)} & 121.1(3.29) & 121.1(3.27) & 119.5(2.99) & 118.9(3.17) & 120.1(3.5) & 118.6(3.28) & 118.9(2.85) & 119.7(3.18) & 118.6(3.26)
\cr\hline
\end{tabular}}
\end{table*}

\begin{table*}[htbp]
\centering
\fontsize{5.7}{13.5}\selectfont
\caption{Hand Gestures}\label{Hand Gestures}
\setlength{\tabcolsep}{2mm}{
\begin{tabular}{|c|c|c|c|c|c|c|c|c|c|c|c|c|}
\hline

Model & Loss & origin & ELBD & Lap & SPEC & MCFS & NDFS & UDFS & Inf & Fisher & RFS & ReliefF \cr\hline
\hline
\multirow{2}{*}{VAE(MF)}     & -ELBO & 161.5(3.12) & \textbf{122.4(4.38)} & 122.9(4.05) & 123.1(4.08) & 137.4(4.97) & 138.1(3.58) & 123.1(4.2) & 123.2(4.55) & 122.9(4.02) & 158.6(3.79) & 123.1(4.17)
\cr\cline{2-13}
                             & L2   & 133.7(3.44) & \textbf{121.4(4.68)} & 122.6(4.13) & 122.8(4.11) & 126.5(4.05) & 126.6(3.18) & 122.9(4.32) & 123.3(4.48) & 123.2(4.51) & 130.6(2.84) & 122.9(4.16) 
\cr\hline
\multirow{2}{*}{VAE(FC)}     & -ELBO & 65.8(1.44) & \textbf{41.7(1.75)} & 107.6(30.77) & 143.1(16.18) & 135.5(27.32) & 84.8(24.86) & 129.4(15.61) & 150.7(23.93) & 43.7(1.28) & 46.7(1.82) & 75.4(11.73)  \cr\cline{2-13}
                             & L2   & 123.5(4.19) & \textbf{116.5(3.57)} &  236.2(63.74) & 306.3(31.73) & 298.3(53.24) & 199.6(52.16) & 286.5(29.99) & 323.7(48.23) & 119.7(3.0) & 125.2(2.34) & 179.9(21.97) 
\cr\hline
\multirow{2}{*}{CVAE(MF)}    & -ELBO & 154.0(2.01) & \textbf{118.6(1.56)} & 119.1(1.29) & 119.2(1.53) & 127.7(4.84) & 135.3(2.34) & 120.5(2.19) & 119.2(1.56) & 119.1(1.5) & 150.4(1.97) & 151.9(2.58) 
\cr\cline{2-13}
                             & L2   & 129.2(1.47) & \textbf{118.4(1.39)} & 118.9(1.72) & 118.9(1.48) & 121.3(2.47) & 122.6(2.01) & 119.0(1.15) & 118.9(1.48) & 118.7(1.24) & 125.7(1.97) & 127.8(2.64)  
\cr\hline
\multirow{2}{*}{CVAE(FC)}    & -ELBO & 61.6(0.44) & \textbf{38.3(0.61)} & 127.8(7.03) & 114.0(13.49) & 123.3(6.57) & 123.8(8.53) & 103.1(24.01) & 112.7(17.12) & 41.1(1.04) & 43.2(0.78) & 100.9(7.49)  \cr\cline{2-13}
                             & L2   & 116.9(0.93) & \textbf{110.5(1.03)} & 281.1(15.32) & 255.2(26.84) & 276.1(14.11) & 273.0(14.43) & 234.6(46.61) & 247.4(34.47) & 115.0(1.81) & 118.6(1.53) & 229.3(15.65)   
\cr\hline
\multirow{2}{*}{NF-VAE(MF)}  & -ELBO & 163.0(1.89) & \textbf{124.1(1.16)} & 124.6(1.18) & 124.6(1.26) & 135.5(10.65) & 138.3(3.65) & 124.2(1.31) & 124.7(1.25) & 124.6(1.02) & 160.5(1.97) & 124.6(0.96)
\cr\cline{2-13}
                             & L2   & 134.1(2.15) & \textbf{123.9(1.25)} & 124.2(1.39) & 123.9(1.16) & 126.3(2.74) & 127.3(2.07) & 124.2(1.27) & 124.3(1.36) & 124.4(1.1) & 133.4(2.09) & 124.0(1.1) 
\cr\hline
\multirow{2}{*}{NF-VAE(FC)}  & -ELBO & 65.5(0.2)  & \textbf{41.0(0.56)} & 130.8(2.14) & 145.7(15.41) & 133.5(10.64) & 86.3(16.38) & 127.3(24.23) & 129.4(2.86) & 43.0(0.19) & 50.0(2.73) & 78.5(18.41)
\cr\cline{2-13}
                             & L2   & 122.3(0.95) & \textbf{115.3(0.98)} & 284.2(6.98) & 316.4(32.14) & 296.0(20.68) & 200.5(33.42) & 278.9(46.28) & 287.3(2.8) & 118.4(0.14) & 130.8(5.31) & 185.4(35.19)  
\cr\hline
\multirow{2}{*}{IAF-VAE(MF)} & -ELBO & 133.0(2.28) & \textbf{126.0(2.69)} & 126.1(2.74) & 126.1(2.68) & 130.5(1.97) & 128.4(2.52) & 126.1(2.7) & \textbf{126.0(2.73)} & 126.5(2.75) & 132.5(2.28) & 126.1(2.62) 
\cr\cline{2-13}
                             & L2   & 128.4(2.69) & \textbf{126.0(2.72)} & \textbf{126.0(2.7)} & 126.1(2.74) & 127.6(2.45) & 126.8(2.67) & 126.1(2.7) & 126.1(2.66) & \textbf{126.0(2.79)} & 128.4(2.67) & 126.1(2.75)  
\cr\hline

\multirow{2}{*}{IAF-VAE(FC)} & -ELBO & 58.7(3.75) & \textbf{41.7(3.63)} & 43.8(4.49) & 42.8(4.81) & 45.3(3.69) & 48.7(4.39) & 43.7(2.61) & 48.9(4.3) & 42.1(3.65) & 41.9(3.57) & 48.3(4.23) 
\cr\cline{2-13}
                             & L2   & 120.5(7.13) & \textbf{118.0(7.22)} & 120.6(8.76) & 119.8(9.25) & 121.0(7.02) & 122.6(8.34) & 119.8(6.49) & 122.3(8.74) & 118.3(7.21) & 118.2(7.25) & 122.0(8.68) 
\cr\hline
\end{tabular}}
\end{table*}

\begin{table*}[htbp]
\centering
\fontsize{6}{13.5}\selectfont
\caption{Fingers}\label{Fingers}
\setlength{\tabcolsep}{2mm}{
\begin{tabular}{|c|c|c|c|c|c|c|c|c|c|c|c|c|}
\hline

Model & Loss & origin & ELBD & Lap & SPEC & MCFS & NDFS & UDFS & Inf & Fisher & RFS & ReliefF \cr\hline
\hline
\multirow{2}{*}{VAE(MF)}     & -ELBO & 52.7(0.9) & 30.0(0.51)  & 30.0(0.49) & \textbf{29.8(0.61)} & 39.3(3.52) & 30.1(0.51) & 31.1(1.57) & 29.8(0.48) & 30.2(0.32) & 37.9(2.24) & 51.6(0.88)  
\cr\cline{2-13}
                             & L2   & 35.2(0.85) & \textbf{29.7(0.48)} & 29.8(0.46) & 29.9(0.37) & 31.7(1.08) & 30.0(0.61) & 29.9(0.51) & 29.7(0.41) & 29.9(0.42) & 31.7(0.69) & 34.6(1.02)
\cr\hline
\multirow{2}{*}{VAE(FC)}     & -ELBO & 14.7(0.08) & \textbf{12.6(0.25)} & 19.1(1.21) & 19.6(0.28) & 17.8(1.39) & 17.8(0.47) & 18.8(0.91) & 17.1(0.81) & 14.0(0.67) & 17.4(1.41) & 18.6(0.46)  \cr\cline{2-13}
                             & L2   & 28.1(0.28) & \textbf{23.9(0.43)} & 36.4(2.52) & 37.9(0.76) & 34.3(2.95) & 34.0(1.12) & 36.3(1.97) & 32.5(1.55) & 26.4(1.29) & 33.3(2.85) & 35.8(0.68)
\cr\hline
\multirow{2}{*}{NF-VAE(MF)}  & -ELBO & 52.1(0.74) & \textbf{29.0(0.11)} & 29.1(0.13) & 29.2(0.2) & 36.3(2.04) & 29.2(0.26) & 30.0(0.85) & \textbf{29.0(0.27)} & 31.3(1.68) & 36.6(3.3) & 51.2(0.54)
\cr\cline{2-13}
                             & L2   & 34.3(0.35) & \textbf{28.9(0.26)} & \textbf{28.9(0.15)} & 29.0(0.23) & 30.1(0.37) & 29.1(0.19) & 29.1(0.13) & \textbf{28.9(0.18)} & 29.3(0.38) & 30.4(0.86) & 33.5(0.18) 
\cr\hline
\multirow{2}{*}{NF-VAE(FC)}  & -ELBO & 14.0(0.06) & \textbf{12.8(0.09)} & 19.3(1.36) & 19.1(1.09) & 18.3(0.47) & 17.7(0.93) & 18.6(1.12) & 17.9(0.99) & 15.5(0.41) & 16.8(0.55) & 19.1(0.25)
\cr\cline{2-13}
                             & L2   & 28.2(0.17) & \textbf{24.2(0.35)} & 37.5(2.67) & 37.1(2.05) & 35.3(1.22) & 34.1(1.79) & 36.0(2.19) & 34.4(1.76) & 29.7(0.57) & 32.3(1.11) & 37.1(0.47) 
\cr\hline
\multirow{2}{*}{IAF-VAE(MF)} & -ELBO & 12.42(0.02) & \textbf{12.35(0.08)} & 12.36(0.08) & 12.36(0.08) & 12.36(0.08) & 12.36(0.08) & 12.36(0.08) & 12.36(0.08) & 12.36(0.08) & \textbf{12.35(0.08)} & 12.36(0.08)
\cr\cline{2-13}
                             & L2   & 12.42(0.02) & \textbf{12.35(0.08)} & \textbf{12.35(0.08)} & \textbf{12.35(0.08)} & \textbf{12.35(0.08)} & \textbf{12.35(0.08)} & \textbf{12.35(0.08)} & \textbf{12.35(0.08)} & \textbf{12.35(0.08)} & \textbf{12.35(0.08)} & 12.36(0.08)
\cr\hline

\multirow{2}{*}{IAF-VAE(FC)} & -ELBO & 7.5(0.96) & \textbf{6.9(0.8)} & 7.2(0.93) & 7.1(0.92) & 7.1(0.96) & 7.1(0.75) & 7.3(0.89) & 7.2(0.75) & 7.0(0.81) & 7.2(0.87) & 7.2(1.03) 
\cr\cline{2-13}
                             & L2   & 26.4(0.91) & \textbf{25.3(0.6)} & 25.7(0.91) & 25.7(0.77) & 25.9(0.92) & 25.6(0.39) & 26.1(0.73) & 25.9(0.64) & 25.4(0.6) & 25.9(0.73) & 25.8(0.87) 
\cr\hline
\end{tabular}}
\end{table*}

\begin{table*}[htbp]
\centering
\fontsize{5.5}{13.5}\selectfont
\caption{Brain}\label{Brain}
\setlength{\tabcolsep}{2mm}{
\begin{tabular}{|c|c|c|c|c|c|c|c|c|c|c|c|c|}
\hline

Model & Loss & origin & ELBD & Lap & SPEC & MCFS & NDFS & UDFS & Inf & Fisher & RFS & ReliefF \cr\hline
\hline
\multirow{2}{*}{VAE(MF)}     & -ELBO & 695.9(8.82) & \textbf{529.2(13.85)} & 529.4(13.51) & 530.1(14.19) & 541.3(15.0) & 562.6(11.19) & 530.8(13.04) & 530.0(13.7) & 550.6(15.56) & 539.8(13.45) & 568.8(18.94) 
\cr\cline{2-13}
                             & L2   & 565.1(13.68) & \textbf{527.5(14.2)} & 528.0(14.21) & 528.9(13.69) & 530.3(13.95) & 533.4(13.59) & 529.6(13.57) & 528.9(13.8) & 531.3(13.99) & 530.4(13.74) & 534.6(14.82) 
\cr\hline
\multirow{2}{*}{VAE(FC)}     & -ELBO & 424.3(9.11) & \textbf{373.9(10.99)} & 414.8(7.02) & 409.6(4.44) & 399.5(15.57) & 400.6(7.64) & 401.6(20.67) & 402.4(9.54) & 400.0(15.22) & 411.7(13.73) & 390.1(13.11) 
\cr\cline{2-13}
                             & L2   & 796.8(19.81) & \textbf{783.4(22.10)} &  861.9(17.05) & 857.0(7.29) & 835.8(31.86) & 840.2(15.49) & 841.1(37.92) & 844.0(20.18) & 836.4(33.03) & 867.0(28.5) & 818.0(26.95)
\cr\hline
\multirow{2}{*}{CVAE(MF)}    & -ELBO & 689.1(8.36) & \textbf{527.1(6.68)} & 527.4(6.87) & 527.6(6.96) & 541.5(3.7) & 567.9(2.74) & 530.0(4.11) & 527.9(6.57) & 548.2(7.35) & 539.1(6.29) & 572.7(4.58) 
\cr\cline{2-13}
                             & L2   & 562.5(7.31) & \textbf{525.1(7.19)} & 526.6(6.92) & 526.4(6.55) & 528.8(6.7) & 533.7(6.47) & 526.5(6.45) & 526.5(6.91) & 529.0(6.81) & 528.3(7.07) & 533.5(5.97) 
\cr\hline
\multirow{2}{*}{CVAE(FC)}    & -ELBO & 413.1(3.04) & \textbf{356.6(1.98)} & 400.7(8.16) & 397.6(3.25) & 380.2(7.55) & 381.1(1.65) & 382.7(3.81) & 400.7(4.86) & 377.0(7.93) & 388.6(6.63) & 383.4(7.35) 
\cr\cline{2-13}
                             & L2   & 771.4(6.94) & \textbf{750.1(4.27)} & 842.5(13.73) & 837.0(3.28) & 797.7(16.8) & 800.7(6.13) & 802.8(6.89) & 840.0(9.97) & 792.9(15.19) & 812.2(12.94) & 807.1(15.95)   
\cr\hline
\multirow{2}{*}{NF-VAE(MF)}  & -ELBO & 694.6(3.33) & \textbf{530.1(3.85)} & 531.1(4.63) & 531.6(3.92) & 543.8(10.38) & 577.3(3.4) & 532.1(4.49) & 531.9(4.94) & 551.7(4.29) & 549.0(3.83) & 571.6(4.86)  
\cr\cline{2-13}
                             & L2   & 567.8(3.4) & \textbf{529.1(4.6)} & 530.4(4.34) & 530.9(4.07) & 532.6(5.43) & 538.2(3.96) & 530.3(4.76) & 530.6(5.18) & 534.4(4.57) & 533.8(3.76) & 537.2(3.81)  
\cr\hline
\multirow{2}{*}{NF-VAE(FC)}  & -ELBO & 421.2(2.87) & \textbf{369.7(2.93)} & 407.7(2.03) & 403.6(5.54) & 389.7(6.6) & 391.3(5.83) & 389.8(3.99) & 406.7(2.44) & 385.3(1.78) & 408.3(2.78) & 383.2(1.07) 
\cr\cline{2-13}
                             & L2   & 789.8(6.22) & \textbf{775.8(4.76)} & 852.9(6.51) & 848.0(18.3) & 817.6(12.16) & 824.3(14.69) & 817.5(6.23) & 852.8(8.91) & 805.8(1.01) & 853.0(6.5) & 806.6(3.11) 
\cr\hline
\multirow{2}{*}{IAF-VAE(MF)} & -ELBO & 626.7(19.02) & 552.1(11.79) & 547.3(17.93) & \textbf{547.2(17.84)} & 550.4(20.03) & 586.1(25.6) & 548.0(18.86) & 547.5(17.68) & 553.5(20.13) & 549.1(19.39) & 575.9(22.71) 
\cr\cline{2-13}
                             & L2   & 562.7(18.11) & \textbf{546.9(17.78)} & 547.5(18.21) & 547.7(18.3) & 548.1(18.46) & 554.6(19.53) & 547.5(18.24) & 547.3(18.22) & 548.6(18.9) & 547.9(18.18) & 553.9(19.48) 
\cr\hline

\multirow{2}{*}{IAF-VAE(FC)} & -ELBO & 367.4(2.38) & \textbf{312.9(2.59)} & 323.8(1.46) & 323.6(0.62) & 319.7(2.42) & 327.7(2.37) & 326.5(1.27) & 338.0(1.48) & 320.5(3.49) & 318.5(1.99) & 313.5(2.56) 
\cr\cline{2-13}
                             & L2   & 704.0(2.66) & \textbf{682.6(3.69)} & 693.2(3.22) & 693.3(2.43) & 689.3(1.08) & 691.3(1.26) & 692.1(2.32) & 699.8(1.93) & 689.4(2.56) & 686.8(3.9) & 684.5(3.89)  
\cr\hline
\end{tabular}}
\end{table*}

\begin{table*}[htbp]
\centering
\fontsize{5.5}{13.5}\selectfont
\caption{Yale}\label{Yale}
\setlength{\tabcolsep}{2mm}{
\begin{tabular}{|c|c|c|c|c|c|c|c|c|c|c|c|c|}
\hline

Model & Loss & origin & ELBD & Lap & SPEC & MCFS & NDFS & UDFS & Inf & Fisher & RFS & ReliefF \cr\hline
\hline
\multirow{2}{*}{VAE(MF)}     & -ELBO & 355.2(3.46) & \textbf{236.2(2.73)} & 285.5(4.83) & 284.3(4.37) & 274.5(8.64) & 239.3(2.39) & 236.6(2.52) & 235.0(1.95) & 299.1(9.88) & 268.5(7.84) & 326.7(3.12)   
\cr\cline{2-13}
                             & L2   & 261.5(5.01) & \textbf{229.6(2.99)} & 239.3(3.73) & 238.2(2.59) & 234.6(3.71) & 229.4(4.97) & 229.3(4.22) & 228.3(3.5) & 238.9(3.27) & 235.0(2.45) & 241.9(2.27)  
\cr\hline
\multirow{2}{*}{VAE(FC)}     & -ELBO & 192.2(10.58) & \textbf{145.9(11.17)} & 178.3(13.23) & 178.7(15.1) & 223.3(9.88) & 173.6(18.51) & 235.4(17.83) & 196.5(8.38) & 159.2(8.56) & 227.2(12.79) & 214.6(25.94)   \cr\cline{2-13}
                             & L2   & 348.3(19.3) & \textbf{334.9(19.47)} & 395.3(26.48) & 390.6(26.63) & 482.7(15.59) & 390.6(34.49) & 521.2(27.56) & 436.2(13.66) & 353.5(18.24) & 487.9(20.11) & 476.7(56.13) 
\cr\hline
\multirow{2}{*}{NF-VAE(MF)}  & -ELBO & 354.8(1.57) & \textbf{232.5(3.41)} & 277.0(5.81) & 283.0(2.47) & 270.8(7.93) & 236.9(3.1) & 235.8(5.25) & 233.1(5.24) & 291.5(4.66) & 271.1(3.8) & 324.7(1.01) 
\cr\cline{2-13}
                             & L2   & 254.3(3.15) & \textbf{222.6(4.14)} & 233.1(5.44) & 234.8(3.44) & 232.5(4.81) & 226.2(3.3) & 227.6(3.53) & 226.3(4.65) & 234.9(4.63) & 234.4(3.54) & 240.9(5.02)
\cr\hline
\multirow{2}{*}{NF-VAE(FC)}  & -ELBO & 172.6(2.34) & 131.1(2.75) & 182.9(10.96) & 168.5(5.74) & 231.5(20.21) & 156.7(1.69) & 210.3(2.06) & 173.6(8.67) & 144.0(1.07) & 237.8(8.15) & 235.0(5.06) 
\cr\cline{2-13}
                             & L2   & 312.5(3.43) & 305.9(6.05) & 403.2(23.45) & 374.0(13.99) & 497.5(38.44) & 358.6(1.86) & 462.7(7.98) & 387.3(18.14) & 324.9(2.21) & 514.0(9.87) & 509.5(6.08)  
\cr\hline
\multirow{2}{*}{IAF-VAE(MF)} & -ELBO & 312.7(8.66) & \textbf{247.7(10.18)} & 282.1(8.69) & 280.2(7.46) & 275.7(9.14) & 251.6(9.68) & 251.6(6.77) & 249.3(10.0) & 292.8(10.18) & 272.3(5.37) & 312.1(9.03)
\cr\cline{2-13}
                             & L2   & 261.7(7.61) & \textbf{247.6(9.77)} & 255.2(10.24) & 255.6(10.08) & 254.1(8.98) & 249.9(9.83) & 250.0(8.8) & 249.4(10.23) & 256.5(9.3) & 253.3(7.66) & 263.2(9.2)
\cr\hline

\multirow{2}{*}{IAF-VAE(FC)} & -ELBO & 158.7(5.24) & \textbf{112.3(3.94)} & 121.1(6.35) & 122.2(5.14) & 124.3(5.63) & 134.1(5.07) & 131.6(6.16) & 133.3(2.36) & 134.6(6.13) & 126.9(7.34) & 117.5(6.88)
\cr\cline{2-13}
                             & L2   & 303.5(9.12) & \textbf{287.7(7.03)} & 296.2(9.29) & 297.2(9.49) & 298.9(9.34) & 305.1(9.8) & 302.1(8.57) & 303.7(9.67) & 304.4(10.91) & 299.9(10.72) & 294.2(11.04)
\cr\hline
\end{tabular}}
\end{table*}

\begin{table*}[htbp]
\centering
\fontsize{5.5}{13.5}\selectfont
\caption{Chest}\label{Chest}
\setlength{\tabcolsep}{2mm}{
\begin{tabular}{|c|c|c|c|c|c|c|c|c|c|c|c|c|}
\hline

Model & Loss & origin & ELBD & Lap & SPEC & MCFS & NDFS & UDFS & Inf & Fisher & RFS & ReliefF \cr\hline
\hline
\multirow{2}{*}{VAE(MF)}     & -ELBO & 642.2(1.6) & \textbf{519.4(3.35)} & 519.7(2.77) & 519.9(3.09) & 538.7(10.61) & 600.4(4.31) & 520.9(3.69) & 520.2(4.02) & 552.2(1.99) & 524.0(2.45) & 580.4(3.84)
\cr\cline{2-13}
                             & L2   & 546.4(3.42) & \textbf{517.8(3.12)} & 518.3(3.12) & 518.7(2.77) & 521.1(3.77) & 532.1(1.27) & 519.0(2.7) & 518.6(2.23) & 525.2(2.27) & 518.9(1.91) & 528.9(2.09)
\cr\hline
\multirow{2}{*}{VAE(FC)}     & -ELBO & 296.7(1.77) & \textbf{249.6(1.61)} & 323.3(4.8) & 319.8(5.78) & 313.7(5.89) & 308.1(14.7) & 318.4(5.9) & 316.8(6.14) & 301.5(9.59) & 314.2(10.67) & 311.1(9.29) 
\cr\cline{2-13}
                             & L2   & 561.0(1.07) & \textbf{549.8(2.67)} &  687.9(3.75) & 682.7(12.94) & 675.4(14.08) & 657.8(23.61) & 683.6(11.87) & 677.4(17.59) & 647.2(19.92) & 673.4(24.64) & 663.8(14.76)
\cr\hline
\multirow{2}{*}{CVAE(MF)}    & -ELBO & 658.9(6.12) & \textbf{536.1(5.26)} & 537.7(5.71) & 535.7(5.04) & 549.7(3.28) & 620.7(8.76) & 537.7(5.87) & 536.8(5.4) & 573.0(3.25) & 543.9(3.83) & 599.6(7.12) 
\cr\cline{2-13}
                             & L2   & 561.0(6.54) & \textbf{534.1(5.43)} & 535.0(5.66) & 536.0(5.14) & 537.6(4.58) & 549.5(5.96) & 535.0(5.2) & 534.8(5.61) & 541.7(5.65) & 535.8(4.86) & 544.5(6.31) 
\cr\hline
\multirow{2}{*}{CVAE(FC)}    & -ELBO & 297.4(1.42) & \textbf{250.7(1.2)} & 314.7(3.94) & 315.2(3.99) & 318.0(7.72) & 303.0(10.62) & 311.8(9.71) & 316.6(3.16) & 310.8(8.2) & 308.9(6.78) & 312.9(2.08) 
\cr\cline{2-13}
                             & L2   & 561.3(2.73) & \textbf{552.1(0.97)} & 668.9(12.21) & 670.7(3.54) & 673.9(17.31) & 651.2(23.32) & 666.6(20.12) & 678.4(10.82) & 667.1(16.86) & 660.1(7.43) & 668.8(3.77)
\cr\hline
\multirow{2}{*}{NF-VAE(MF)}  & -ELBO & 646.2(3.57) & \textbf{523.3(3.09)} & 524.1(2.45) & 525.1(3.17) & 544.5(4.98) & 611.3(2.04) & 525.3(2.8) & 523.9(3.48) & 559.9(7.96) & 530.0(4.34) & 578.7(8.35)  
\cr\cline{2-13}
                             & L2   & 547.1(3.63) & \textbf{522.5(3.20)} & 523.3(3.03) & \textbf{522.5(3.51)} & 525.5(2.6) & 535.9(3.11) & \textbf{522.5(2.96)} & 522.7(2.98) & 527.7(5.0) & 523.7(2.99) & 531.2(3.05)  
\cr\hline
\multirow{2}{*}{NF-VAE(FC)}  & -ELBO & 270.3(0.86) & \textbf{228.0(1.79)} & 272.5(6.96) & 270.8(3.51) & 278.9(3.04) & 256.4(4.7) & 271.4(2.65) & 277.3(3.23) & 272.4(4.07) & 281.7(1.81) & 277.6(0.56) 
\cr\cline{2-13}
                             & L2   & 507.9(2.76) & \textbf{506.3(1.74)} & 585.0(10.71) & 581.9(7.36) & 602.8(10.68) & 554.9(7.48) & 587.2(2.58) & 602.8(5.41) & 587.3(5.97) & 604.4(2.96) & 596.6(3.7) 
\cr\hline
\multirow{2}{*}{IAF-VAE(MF)} & -ELBO & 652.0(9.45) & \textbf{588.7(5.33)} & 591.6(5.23) & 591.6(5.51) & 600.7(8.95) & 646.5(11.27) & 591.3(5.67) & 591.8(5.49) & 616.8(4.68) & 593.9(5.47) & 633.8(4.55)
\cr\cline{2-13}
                             & L2   & 600.0(5.95) & \textbf{588.6(5.35)} & 591.2(5.41) & 591.8(5.18) & 592.1(5.98) & 599.8(5.95) & 591.5(5.28) & 591.3(5.31) & 594.9(5.37) & 591.9(5.4) & 597.4(5.41)
\cr\hline

\multirow{2}{*}{IAF-VAE(FC)} & -ELBO & 333.4(1.67) & \textbf{287.2(2.89)} & 295.3(2.93) & 292.0(2.94) & 304.5(2.92) & 310.3(2.46) & 302.5(2.5) & 312.2(1.2) & 294.4(1.46) & 295.3(0.24) & 295.3(2.23) 
\cr\cline{2-13}
                             & L2   & 651.7(5.03) & \textbf{642.7(5.64)} & 651.4(2.44) & 649.5(4.28) & 653.7(3.89) & 653.1(4.01) & 651.3(4.31) & 655.3(3.73) & 649.2(4.21) & 650.4(2.98) & 651.0(2.46)  
\cr\hline
\end{tabular}}
\end{table*}

\begin{table*}[htbp]
\centering
\fontsize{10}{15}\selectfont
\caption{Basic information about data sets of classification tasks}\label{classification information}
\setlength{\tabcolsep}{3mm}{
\begin{tabular}{|c|c|c|c|c|}
\hline
\multirow{2}{*}{Data set} & Number of  & Number of pixels &  \multirow{2}{*}{Categories}     & Training set size vs   \cr   
                          & pictures   & on one picture   &                 & Testing set size   
\cr\hline 
\hline
Fashion-MNIST          & 70000    & 28$\times$28            & 10  & 6:1  \cr\hline
EMNIST                 & 1456000  & 28$\times$28            & 26  & 6:1  \cr\hline
Shape                  & 27292    & 32$\times$32            & 4   & 4:1  \cr\hline
Chinese Calligraphy    & 105029   & 32$\times$32            & 20  & 4:1  \cr\hline
Eyes                   & 11525    & 3$\times$28$\times$28   & 2   & 4:1  \cr\hline
\end{tabular}}
\end{table*}

\begin{table*}[htbp]
\centering
\fontsize{6.5}{18}\selectfont
\caption{Accuracy on data set of classification tasks}\label{classification}
\setlength{\tabcolsep}{2mm}{
\begin{tabular}{|c|c|c|c|c|c|c|c|c|c|c|}
\hline

Data set & gELBD & Lap & SPEC & MCFS & NDFS & UDFS & Inf & Fisher & RFS & ReliefF \cr\hline
\hline
Fashion-MNIST       &  \textbf{0.872(0.001)} & 0.870(0.002) & 0.869(0.002) & 0.836(0.004) & 0.838(0.001) & 0.869(0.005) & 0.834(0.002) & \textbf{0.872(0.002)} & 0.867(0.004) & 0.867(0.002) \cr\hline
EMNIST              &  \textbf{0.909(0.002)} & 0.835(0.004) & 0.861(0.001) & 0.852(0.002) & 0.905(0.002) & 0.905(0.001) & 0.853(0.003) & 0.908(0.002) & 0.893(0.003) & 0.907(0.001) \cr\hline
Shape               &  \textbf{0.883(0.005)} & 0.819(0.008) & 0.821(0.008) & 0.848(0.010) & 0.812(0.004) & 0.882(0.004) & 0.840(0.012) & 0.864(0.005) & 0.863(0.010) & 0.865(0.011) \cr\hline
Chinese Calligraphy &  \textbf{0.806(0.003)} & 0.745(0.001) & 0.785(0.007) & 0.748(0.001) & 0.799(0.002) & 0.795(0.001) & 0.787(0.002) & 0.800(0.003) & 0.786(0.005) & 0.779(0.005) \cr\hline
Eyes                &  \textbf{0.860(0.008)} & 0.839(0.016) & 0.847(0.012) & 0.838(0.006) & 0.843(0.008) & 0.852(0.005) & 0.851(0.017) & 0.842(0.003) & 0.845(0.006) & 0.846(0.011) \cr\hline

\end{tabular}}
\end{table*}

\begin{figure*}[htbp]
\centering

\subfigure[]
{
    \begin{minipage}[b]{.48\linewidth}
        \centering
        \includegraphics[scale=0.65]{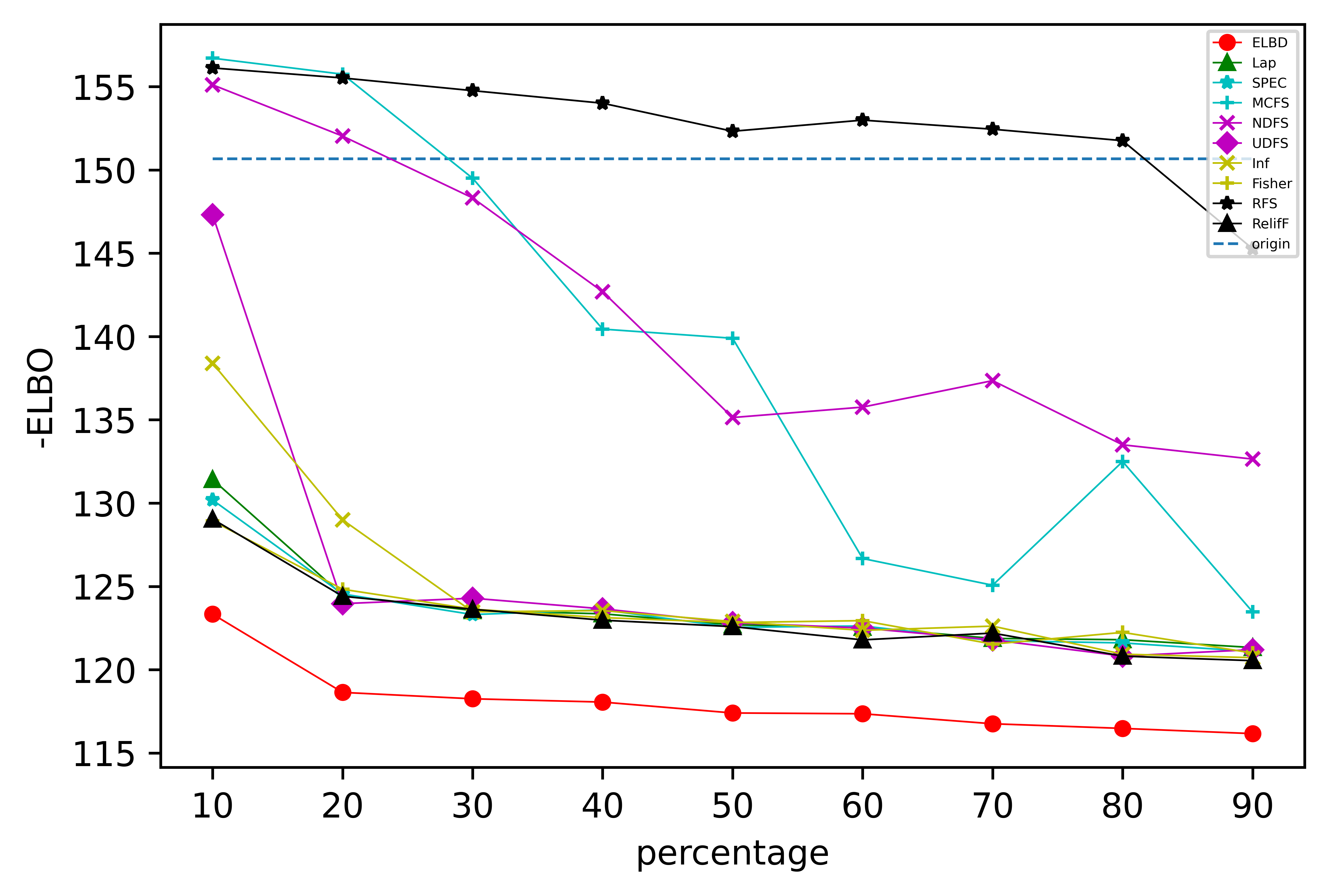}
    \end{minipage}
}
\subfigure[]
{
    \begin{minipage}[b]{.48\linewidth}
        \centering
        \includegraphics[scale=0.65]{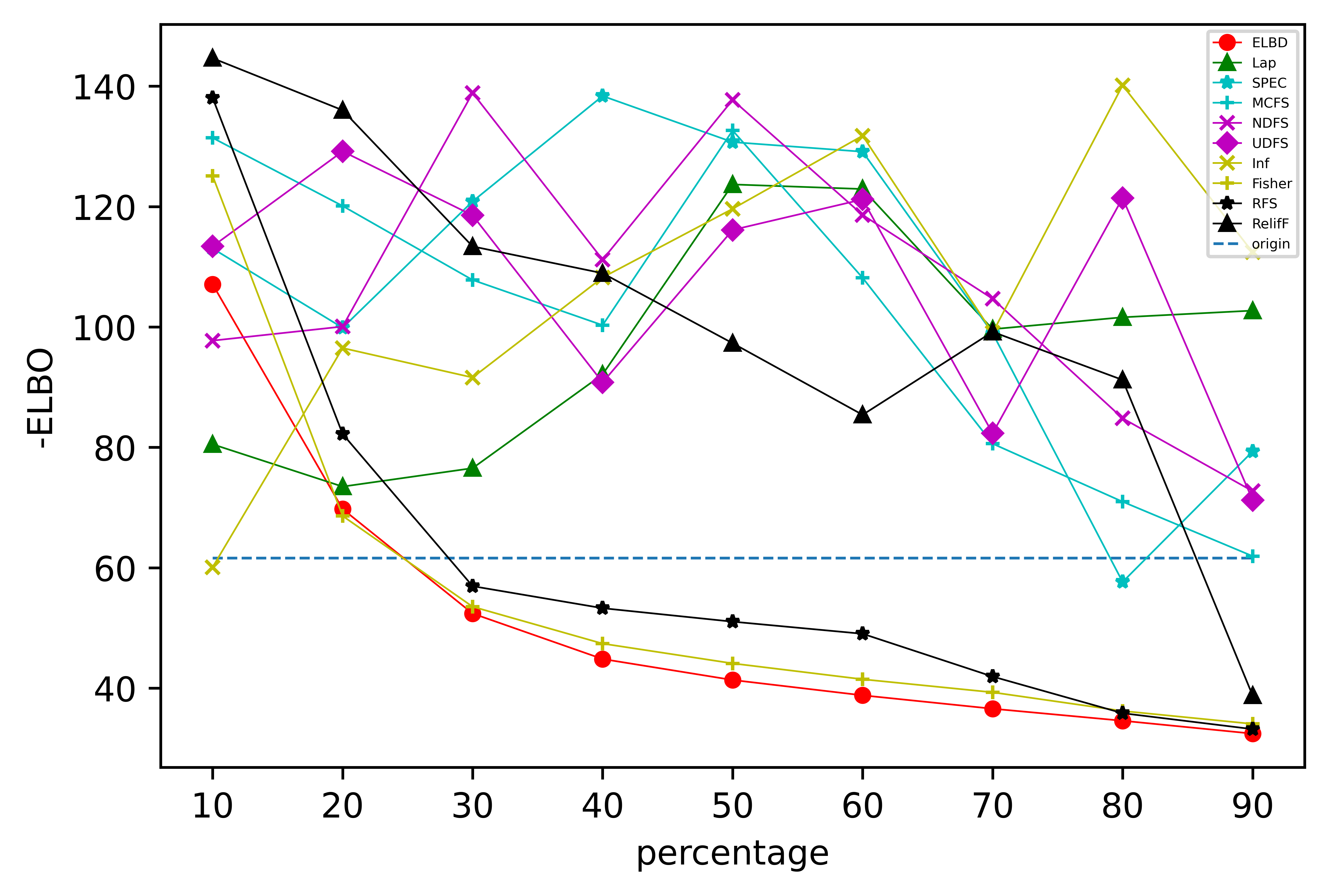}
    \end{minipage}
}
\subfigure[]
{
 	\begin{minipage}[b]{.48\linewidth}
        \centering
        \includegraphics[scale=0.65]{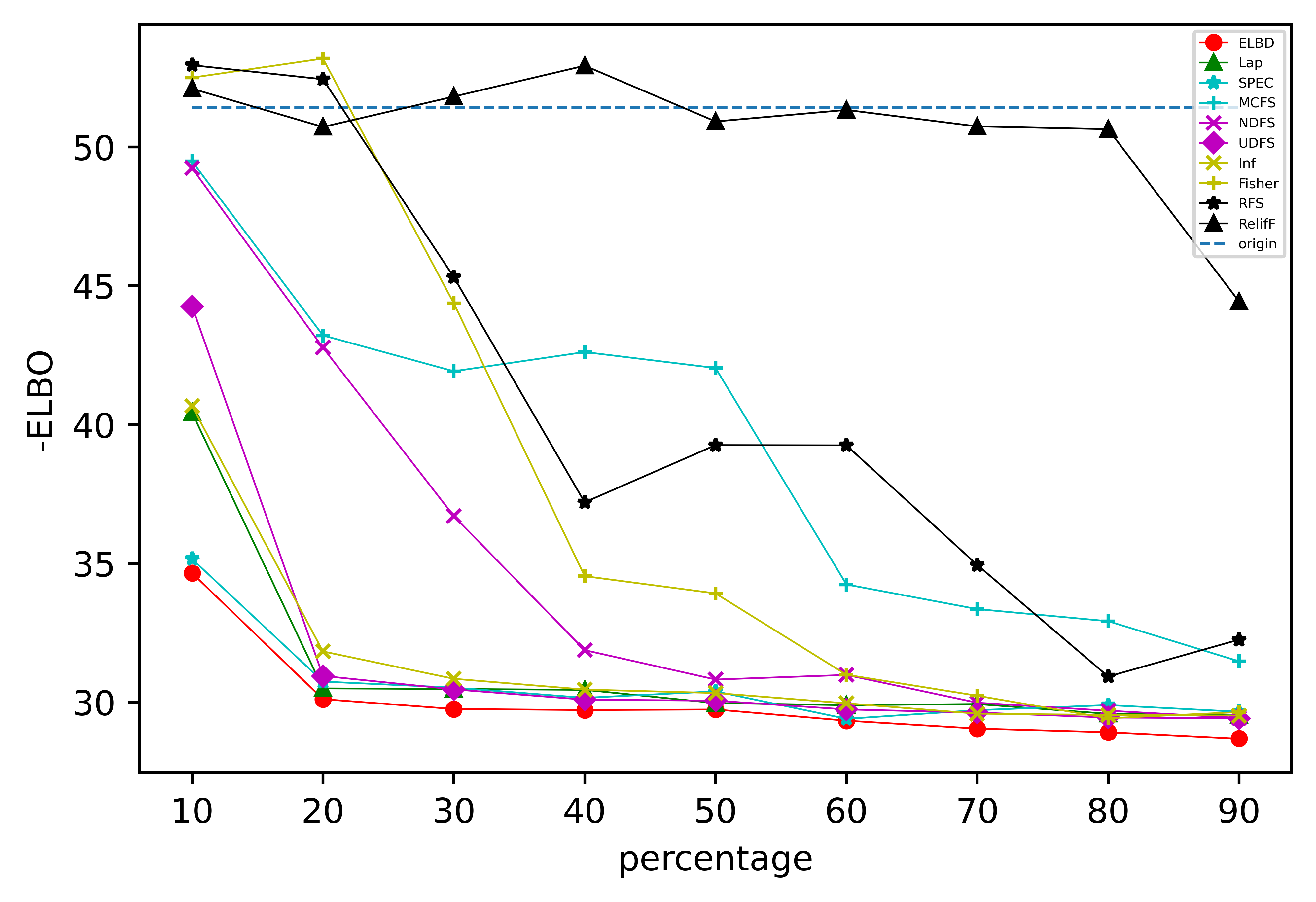}
    \end{minipage}
}
\subfigure[]
{
 	\begin{minipage}[b]{.48\linewidth}
        \centering
        \includegraphics[scale=0.65]{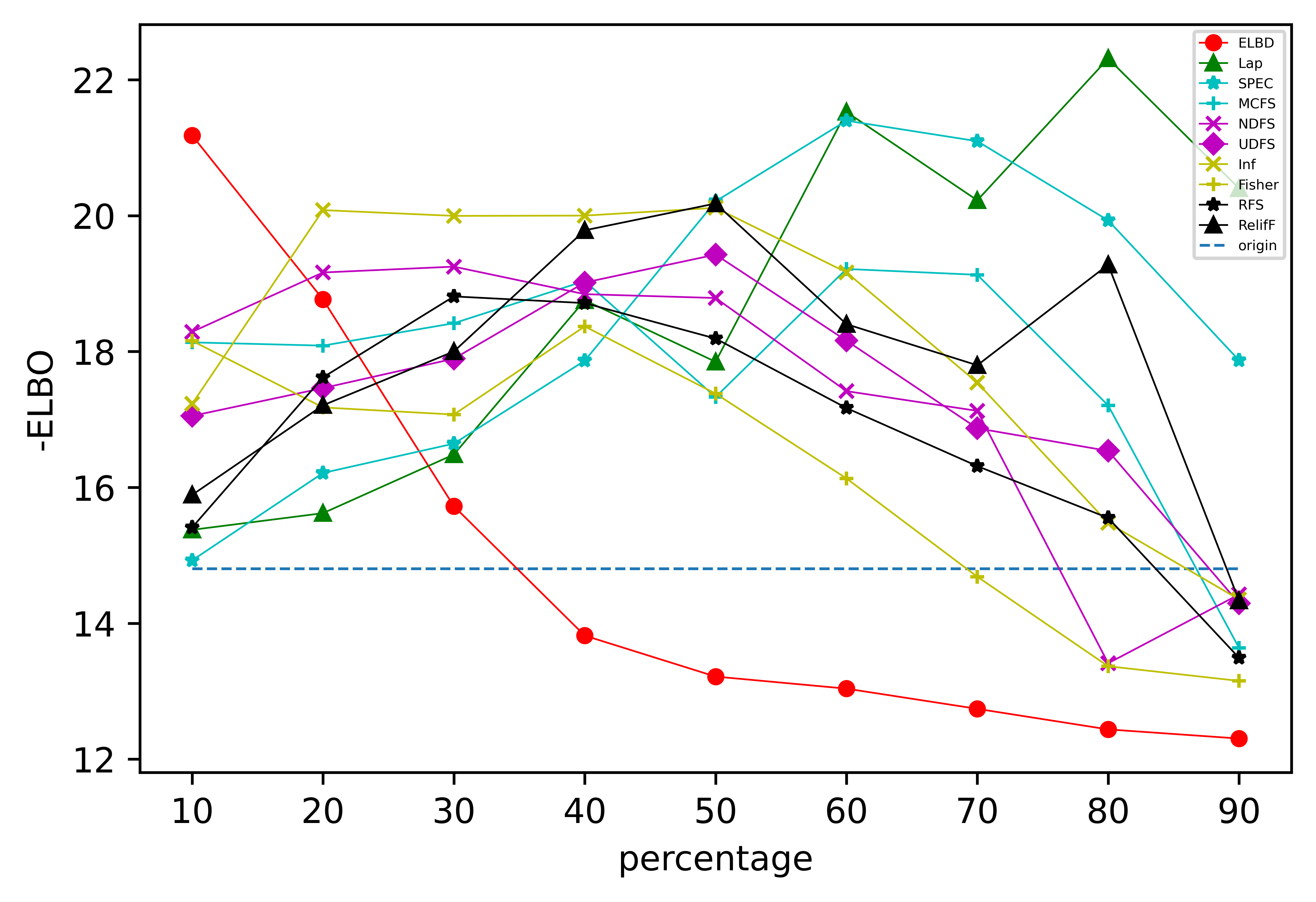}
    \end{minipage}
}    

\subfigure[]
{
 	\begin{minipage}[b]{.48\linewidth}
        \centering
        \includegraphics[scale=0.65]{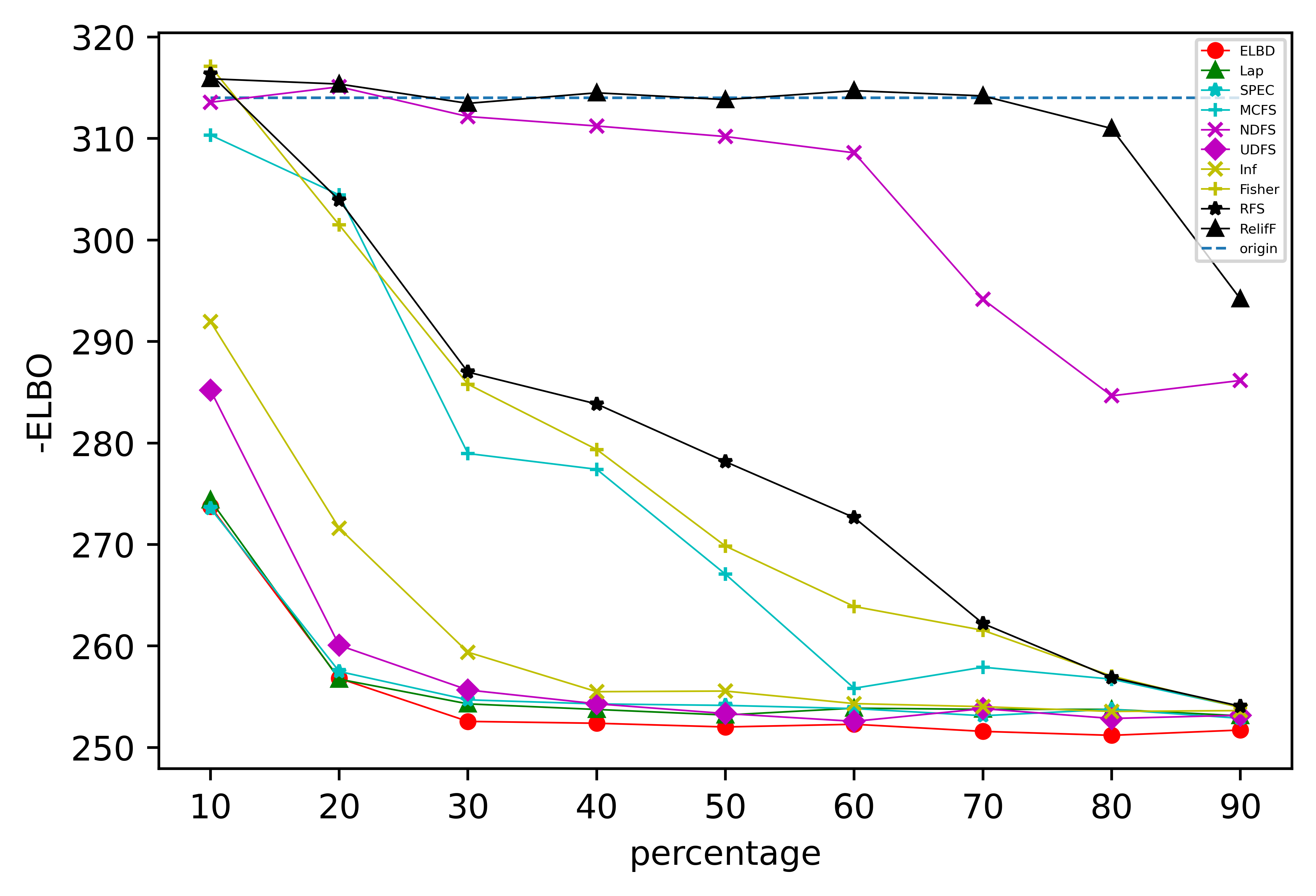}
    \end{minipage}
}
\subfigure[]
{
 	\begin{minipage}[b]{.48\linewidth}
        \centering
        \includegraphics[scale=0.65]{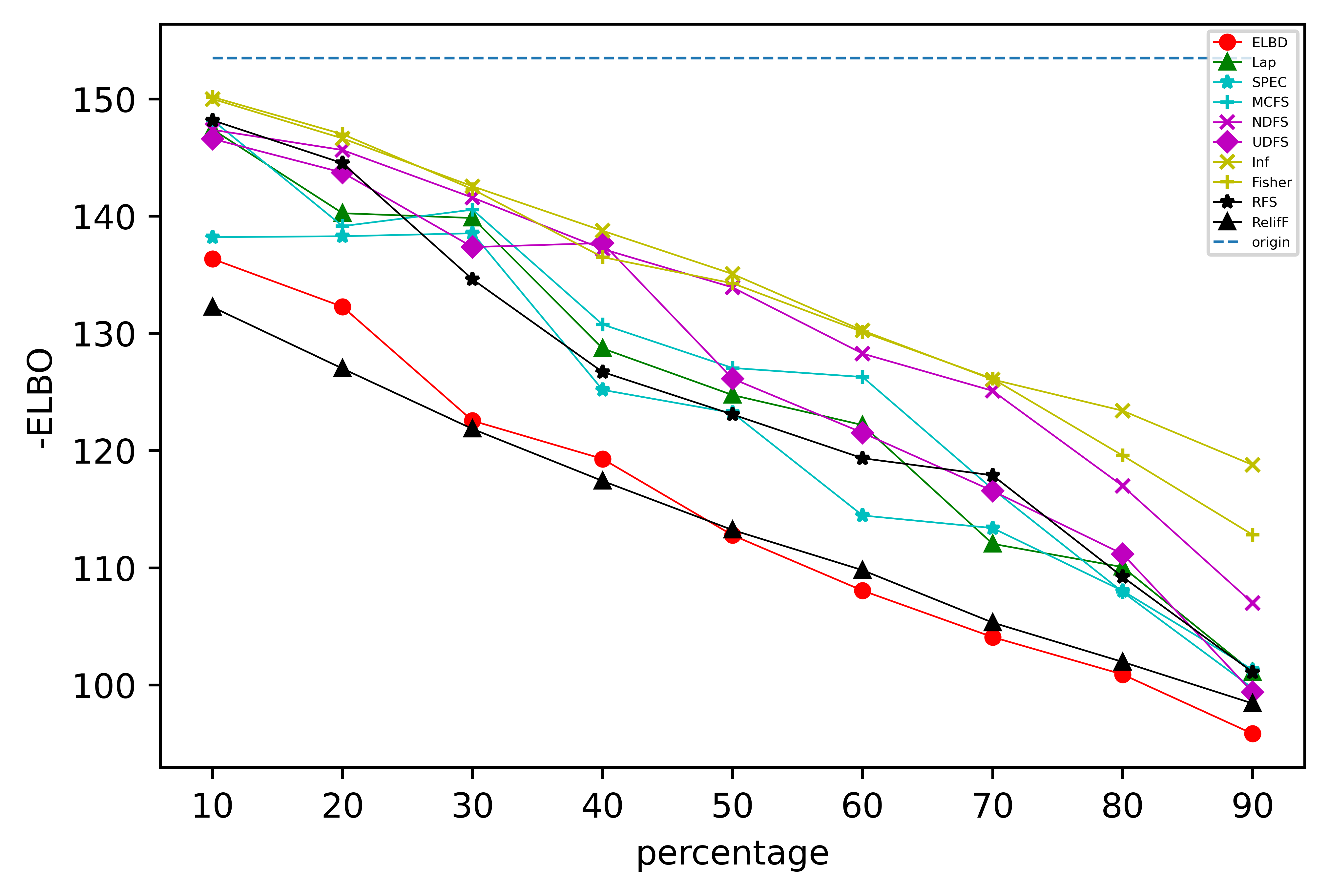}
    \end{minipage}
}    
\caption{(a) The CVAE model on Hand Gestures data set under mean-field condition (b) The CVAE model on Hand Gestures data set under full-covariance condition (c) The NF-VAE model on Fingers data set under mean-field condition (d) The NF-VAE model on Finger data set under full-covariance condition (e) The IAF-VAE model on Yale data set under mean-field condition (f) The IAF-VAE model on Yale data set under full-covariance condition.}
\end{figure*}

In this subsection, we mainly use the marginalization of encoder algorithms and the weak convergence approximation algorithms in Section 3 to optimize VAE and its variants. 

Firstly, we give some information about data sets and part of hyperparameters used in these data sets in Table \ref{data information}. The ``Categories'' term means how many types of labels in the data sets where Fingers and Yale do not have labels. We split each data set into train set and test set randomly. The ratios are shown in ``Train set vs Test set'' term. The ``Dimension of latent variables'' term means the dimension of $\mathbf z$ used on all the models under this data set, i.e., latent variables of VAE and variants have the same dimension under one data set. Considering the capacity of memory and complexity, we choose a mini-batch of data set to compute feature selection scores for latent variables. The sizes of the mini-batch are listed in term ``Mini-batch size''. We also use the same size of mini-batch on all the feature selection algorithms in one data set. The term ``$K$'' is the hyperparameter in \textbf{Algorithm \ref{approximation P mf}} and \textbf{Algorithm \ref{approximation P fc}}. We choose ``Mini-batch size'' and ``$K$'' as much large as we can because of \textbf{Corollary \ref{loss convergence}}.

Data sets with labels, MNIST, KMNIST, Hand Gestures, Chest, Brain, are used to test the algorithms on VAE, CVAE, NF-VAE and IAF-VAE each under two types of latent variables totally 8 models. Because CVAE is a supervised model, data sets without labels, Fingers and Yale, are tested only on VAE, NF-VAE and IAF-VAE each under two types of latent variables totally 6 models. We use full connect neural networks on data sets MNIST, KMNIST, and convolutional neural networks on data sets Hand Gestures, Fingers, Brain, Yale and Chest to build the encoder networks and decoder networks. We use negative ELBO as the loss function for all models and use Adam algorithm\cite{kingma2014adam} to minimize the loss. The learning rate is in $[1\times10^{-3},1\times10^{-5}]$. 

Considering the marginalization of encoder is rather help to generate pictures but give a mathematical explanation of ELBO after optimization, we use negative ELBO and L2 two kinds of losses to measure the results of models on test data set but only use negative ELBO in training steps. The decreasing of both losses can state that our algorithms do generate much more clear pictures

The whole experiments run as follow: Firstly, we train the VAE or its variants without feature selection, and obtain a trained original model $M$. Secondly, for ELBD, we directly use \textbf{Algorithm \ref{ELBD mean-field}} or \textbf{Algorithm \ref{ELBD full-covariance}} to compute the score of every latent variable. For other feature selection algorithms, we turn data set $\{\mathbf x_1,...,\mathbf x_N\}$ to set of latent variables $\{\mathbf z_1,...,\mathbf z_N\}$ by using encoder in $M$. We see the $\{\mathbf z_1,...,\mathbf z_N\}$ as new data set and use other 9 algorithms on it to obtain the scores for every latent variable. Thirdly, we pick the $60\%$ of latent variables which have the largest feature selection scores and see them as $\mathbf u$. Then we build $\boldsymbol\pi$ by the rule that $\pi_i$ is zero if $i$ is one of the indices that we pick and is one otherwise. Different feature selection algorithm generates different $\mathbf u$ and $\boldsymbol\pi$. Finally, we put $\boldsymbol\pi$, $M$ and $K$ into \textbf{Algorithm \ref{marginal mf Q}}, \textbf{Algorithm \ref{approximation P mf}} if latent variables of $M$ have mean-field Gaussian posterior or \textbf{Algorithm \ref{marginal fc Q}}, \textbf{Algorithm \ref{approximation P fc}} if latent variables of $M$ have full-covariance Gaussian posterior to optimize $M$. Using the new pictures generated by the weak convergence algorithms, we can compute the first term of ELBO (\ref{ELBO}). Combing the $\mathcal{KL}$ divergence from \textbf{Algorithm \ref{marginal mf Q}} or \textbf{Algorithm \ref{marginal fc Q}}, we have the ELBO of model after optimization. Combing the generated pictures and original pictures, we have the L2 loss of model after optimization.

Table 2 to Table 8 are the results on 7 data sets in Table \ref{data information}. The titles of them state the data set we run our experiments on. In each table, MF stands for mean-field and FC stands for full-covariance. For example, VAE(MF) means VAE model with mean-field Gaussian posterior. Every column uses one feature selection algorithm to generate $\boldsymbol\pi$ for different models. Every row uses different feature selection algorithms to compute $\boldsymbol\pi$ for one model where ``origin'' term shows the results of original models without any optimization algorithms. We train three models with different random seeds in each kind of model in each data set. We list mean and standard deviation of the three models as the form that mean(standard deviation) in each entry. The lowest result in each row is bold.

Analyzing the Table 2 to Table 8, we find that most of the feature selection algorithms can optimize the models compare to the results in ``origin'' and all the models can be optimized. Moreover, in most of the cases, the optimized results are tremendous. For example, in Yale data set Table \ref{Yale}, the optimized result of VAE(MF) by ELBD is $33.5\%$ less than original result with respect to -ELBO loss and $12.2\%$ with respect to L2 loss. The optimized result (236.2) is even smaller than the result of original state-of-the-art model IAF-VAE(MF) (312.7). It states that our optimization algorithms in Section 3 are very effective and can be used on many variants of VAE. Compared to other 9 feature selection algorithms, ELBD has the best optimized results in almost any cases. Some algorithms can not even have optimized effect. For example, CVAE(FC) optimized by ReliefF in MNIST data set has -ELBO loss (36.9) which is bigger than the original result (29.2). Recalling the analysis in the end of Section 3, i.e., the more important selected latent variables are, the better optimized result is, we also can derive that ELBD is better in picking important latent variables than other 9 algorithms.

\begin{figure*}[htbp]
\centering

\subfigure[]
{
    \begin{minipage}[b]{.3\linewidth}
        \centering
        \includegraphics[scale=0.8]{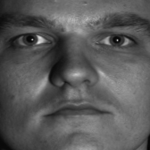}
    \end{minipage}
}
\subfigure[]
{
    \begin{minipage}[b]{.3\linewidth}
        \centering
        \includegraphics[scale=0.8]{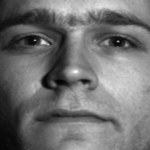}
    \end{minipage}
}
\subfigure[]
{
 	\begin{minipage}[b]{.3\linewidth}
        \centering
        \includegraphics[scale=0.8]{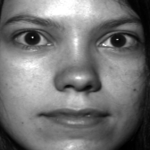}
    \end{minipage}
}

\subfigure[]
{
    \begin{minipage}[b]{.3\linewidth}
        \centering
        \includegraphics[scale=0.8]{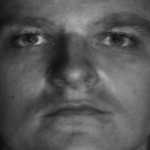}
    \end{minipage}
}
\subfigure[]
{
    \begin{minipage}[b]{.3\linewidth}
        \centering
        \includegraphics[scale=0.8]{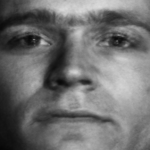}
    \end{minipage}
}
\subfigure[]
{
 	\begin{minipage}[b]{.3\linewidth}
        \centering
        \includegraphics[scale=0.8]{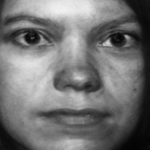}
    \end{minipage}
}

\subfigure[]
{
    \begin{minipage}[b]{.3\linewidth}
        \centering
        \includegraphics[scale=0.8]{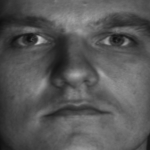}
    \end{minipage}
}
\subfigure[]
{
    \begin{minipage}[b]{.3\linewidth}
        \centering
        \includegraphics[scale=0.8]{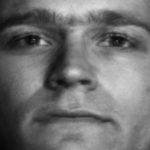}
    \end{minipage}
}
\subfigure[]
{
 	\begin{minipage}[b]{.3\linewidth}
        \centering
        \includegraphics[scale=0.8]{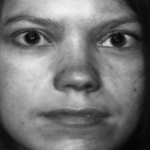}
    \end{minipage}
}
\caption{(a) (b) (c) are original pictures in Yale data set. (d) (e) (f) are pictures generated by VAE(MF). (g) (h) (i) are pictures generated by VAE(MF) optimized by the weak convergence approximation algorithm under ELBD.}
\end{figure*}

Constrained by the computational complexity, we only pick $60\%$ of latent variables to optimize models. However, we pick six different models to study the impact of the percentage on optimized result. The studies are shown in Figure 4. In each graph of Figure 4, the y axis represents -ELBO loss and x axis represents the percentage of latent variables we take by all feature selection algorithms from $10\%$ to $90\%$ (we can not pick $100\%$ since all the analyses in Section 3 are based on the assumption that $\dim\mathbf w>0$ and $\dim\mathbf u>0$). We find that for ELBD, in most of graphs, losses decrease with increasing of percentage of selected latent variables, and become stable or unchanged after certain percentage. It means the latent variable which has low ELBD score basically have no contribution on optimization. But for other feature selection algorithms, things become different. For example, models optimized by Inf in graph (b) and Lap in (d) become worse with increasing of percentage. We also find some algorithms can not optimize models. Therefore, not all the latent variables are worth being chosen as $\mathbf u$ and taking too much ``bad'' latent variables can increase the loss. However in graph (f), all the algorithms show strictly decreasing lines, we can draw the conclusion that all the latent variables in this latent variables of model are ``good''.

In Figure 5, we also show the original pictures on test set of Yale, pictures generated by model and pictures generated by optimized model. We may find that model after optimization generates more clear pictures.

\subsection{gELBD Score on Classification task}

In this subsection, we compare gELBD score with other 9 feature selection algorithms on classification tasks. We use full connect neural network to train the classification models. We use cross entropy as loss function in training step and use Adam algorithm to minimize the loss function. The learning rate is in $[1\times10^{-3}, 1\times10^{-4}]$.

As we mentioned at the end of Section 4, we need to discretize the features before applying gELBD. We multiply 255 and round-down to the closest integer on each feature. Then the value of every feature becomes the integer in $[0,255]$.

We still pick $60\%$ of features and use these features to train models. After getting selected features, we train five different models with different random seeds. The mean and standard deviation of accuracy $\frac RN$ is shown in Table \ref{classification} where $R$ is the number of instances which are classified correctly and $N$ is size of data set.

We can find that gELBD has the highest accuracy in these five data sets compare to other feature selection algorithms.

\section{Conclusion}

This paper is motivated by the fact that eliminating latent variables in VAE can increase the ELBO. According to this fact, this paper proposes algorithms to marginalize the encoder and proposes the weak convergence approximation algorithms to approximate decoder after eliminating latent variables $\mathbf u$. We also do lots of theoretical analyses to guarantee the convergence. We find that the weak convergence approximation algorithms actually increase the weights of distribution $Q(\mathbf u|\mathbf x)$. Therefore we need to select important latent variables and do the weak convergence approximation algorithms only on them. Then we analyze the necessity of selecting latent variables and present a new feature selection algorithm ELBD and generalize ELBD on classification tasks as gELBD. In experiments, we use two kinds of loss functions to measure the optimized result on test sets. We find that both weak convergence approximation algorithms and ELBD have the best performance. And gELBD also achieves the highest accuracy in classification tasks. Besides VAE, CVAE, NF-VAE, IAF-VAE, we can use our algorithms on other variants of VAE. In latter research, we aim to expand the weak convergence approximation algorithms on more variational inference models. The computational complexity is the fatal drawback of ELBD and gELBD. How to overcome this drawback also needs our exploration.

\ifCLASSOPTIONcompsoc
  % The Computer Society usually uses the plural form
  \section*{Acknowledgments}
\else
  % regular IEEE prefers the singular form
  \section*{Acknowledgment}
\fi
This work was funded by the National Nature Science Foundation of China under Grant 12071428 and 62111530247, and the Zhejiang Provincial Natural Science Foundation of China under Grant LZ20A010002.

% Can use something like this to put references on a page
% by themselves when using endfloat and the captionsoff option.
\ifCLASSOPTIONcaptionsoff
  \newpage
\fi

\bibliographystyle{IEEEtran}
\bibliography{references}

\newpage
\begin{figure}
\centering
\includegraphics[scale=1.5]{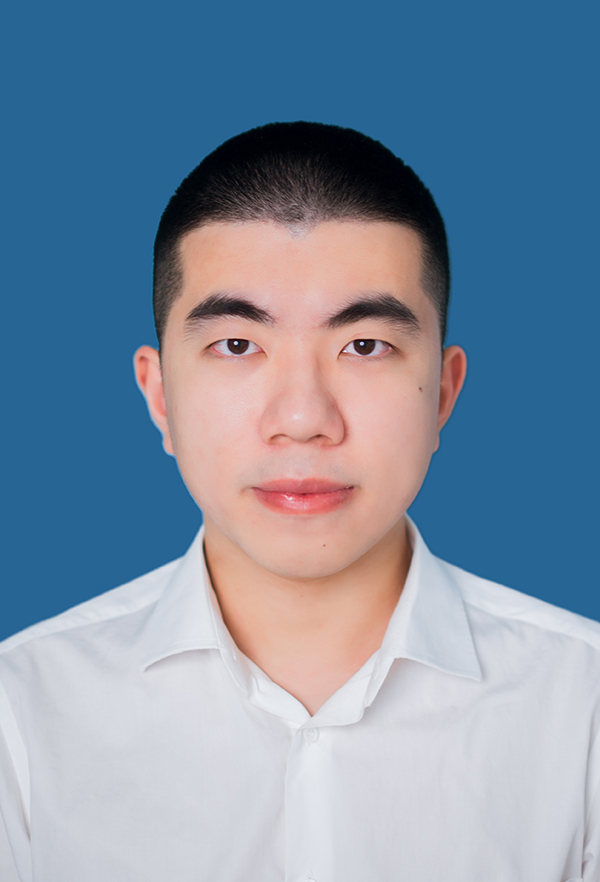}
\end{figure}

Yiran Dong received the B.S. degree in Mathematics and Applied Mathematics from Sichuan University, China, in 2020. He is currently studying towards the Ph.D. degree in operational research and cybernetics in Zhejiang University. His researches mainly focus on generative models, Bayesian network and causal inference.

\begin{figure}[H]
\centering
\includegraphics[scale=0.2]{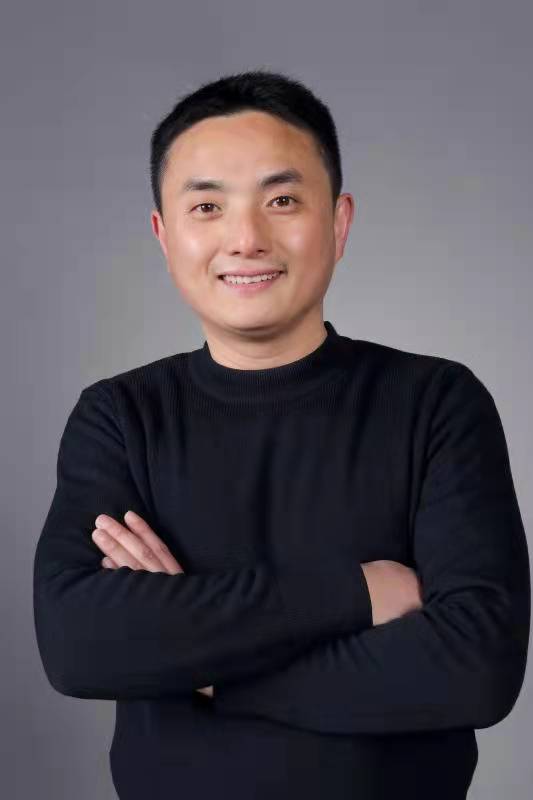}
\end{figure}

Chuanhou Gao received the B.Sc. degree in Chemical Engineering from Zhejiang University of Technology, China, in 1998, and the Ph.D. degree in Operational Research and Cybernetics from Zhejiang University, China, in 2004. From June 2004 until May 2006, he was a Postdoctor in the Department of Control Science and Engineering at Zhejiang University. Since June 2006, he has joined the Department of Mathematics at Zhejiang University, where he is currently a full Professor. He was a visiting scholar at Carnegie Mellon University from Oct. 2011 to Oct. 2012. 
His research interests are in the areas of data-driven modeling, control and optimization, chemical reaction network theory and thermodynamic process control. He is an associate editor of IEEE Transactions on Automatic Control and of International Journal of Adaptive Control and Signal Processing.
\end{document}

\IEEEdisplaynontitleabstractindextext
\IEEEpeerreviewmaketitle
\ifCLASSOPTIONcompsoc
\else

\fi
\appendices
\section{Proof}
\begin{proof}[\textbf{Proof of Proposition 1}]
Let $\mathbf z=(\mathbf w,\mathbf u)$. We already know that 

\begin{align*}
    &\mathcal{KL}[Q(\mathbf z|\mathbf x)\parallel P(\mathbf z|\mathbf x)]-\mathcal{KL}[Q(\mathbf w|\mathbf x)\parallel P(\mathbf w|\mathbf x)]\\=
    &E_{Q(\mathbf w|\mathbf x)}\left(\mathcal{KL}[Q(\mathbf u|\mathbf w,\mathbf x)\parallel P(\mathbf u|\mathbf w,\mathbf x)]\right)\geq 0.
\end{align*}

Then $\mathcal{KL}[Q(\mathbf z|\mathbf x)\parallel P(\mathbf z|\mathbf x)]\geq\mathcal{KL}[Q(\mathbf w|\mathbf x)\parallel P(\mathbf w|\mathbf x)]$, by the definition of ELBO

\begin{align*} 
&\displaystyle\log P(\mathbf x)-\mathcal{KL}[Q(\mathbf z|\mathbf x)\parallel P(\mathbf z|\mathbf x)]\\
=&E_{Q(\mathbf z|\mathbf x)}\left(\log P(\mathbf x|\mathbf z)\right)-\mathcal{KL}[Q(\mathbf z|\mathbf x)\parallel P(\mathbf z)],\notag
\end{align*}

We have the relationship of these two ELBO: 
\begin{align*}
E_{Q(\mathbf z|\mathbf x)}\left(\log P(\mathbf x|\mathbf z)\right)-\mathcal{KL}[Q(\mathbf z|\mathbf x)\parallel P(\mathbf z)]\\ \leq E_{Q(\mathbf w|\mathbf x)}\left(\log P(\mathbf x|\mathbf w)\right)-\mathcal{KL}[Q(\mathbf w|\mathbf x)\parallel P(\mathbf w)].
\end{align*}
By the inference process of ELBO, we know $Q(\mathbf w|\mathbf x)=\int^{+\infty}_{-\infty}Q(\mathbf z|\mathbf x)d\mathbf u$ and $\int^{+\infty}_{-\infty}P(\mathbf x|\mathbf z)Q(\mathbf z|\mathbf x)/P(\mathbf x)d\mathbf u=P(\mathbf w|\mathbf x)$. Then
\begin{align*}
\int^{+\infty}_{-\infty}P(\mathbf x|\mathbf z)Q(\mathbf z|\mathbf x)/Q(\mathbf w|\mathbf x)d\mathbf u=P(\mathbf w|\mathbf x)P(\mathbf x)/Q(\mathbf w|\mathbf x)=P(\mathbf x|\mathbf w).
\end{align*}
So $\int^{+\infty}_{-\infty} P(\mathbf x|\mathbf z)Q(\mathbf u|\mathbf w,\mathbf x)d\mathbf u=P(\mathbf x|\mathbf w)$. This completes the proof.
\end{proof}

\begin{proof}[\textbf{proof of Theorem 1}]
Let $P_K(\mathbf x|\mathbf w)=\mathcal N(\boldsymbol\mu_K,\mathbf I)$. By the definition of $\mathcal{KL}$ divergence,
\begin{align*}
\mathcal{KL}[M_K(\mathbf x|\mathbf w)\parallel P_K(\mathbf x|\mathbf w)]=\int^{+\infty}_{-\infty}M_K(\mathbf x|\mathbf w)\log\frac{M_K(\mathbf x|\mathbf w)}{P_K(\mathbf x|\mathbf w)}d\mathbf x.
\end{align*}
Now we need to find the minimum point of this $\mathcal{KL}$ divergence related to $\boldsymbol\mu_K$, so we take the derivative of $\mathcal{KL}[M_K(\mathbf x|\mathbf w)\parallel P_K(\mathbf x|\mathbf w)]$ with respect to $\boldsymbol\mu_K$.
\begin{align*}
&\frac{d}{d\boldsymbol\mu_K}\mathcal{KL}[M_K(\mathbf x|\mathbf w)\parallel P_K(\mathbf x|\mathbf w)]\\
=&\frac{d}{d\boldsymbol\mu_K}\int^{+\infty}_{-\infty}M_K(\mathbf x|\mathbf w)\log\frac{M_K(\mathbf x|\mathbf w)}{P_K(\mathbf x|\mathbf w)}d\mathbf x\\
=&-\frac{d}{d\boldsymbol\mu_K}\int^{+\infty}_{-\infty}\frac1K\sum^K_{j=1}\mathcal N(\mathbf f(\mathbf w,\mathbf u_j),\mathbf I)\log\exp(-\frac12(\mathbf x-\boldsymbol\mu_K)^2)d\mathbf x\\
=&\frac{d}{d\boldsymbol\mu_K}\int^{+\infty}_{-\infty}\frac1K\sum^K_{j=1}\mathcal N(\mathbf f(\mathbf w,\mathbf u_j),\mathbf I)(\boldsymbol\mu_K-\mathbf x)^2d\mathbf x\\
=&\int^{+\infty}_{-\infty}\frac1K\sum^K_{j=1}\mathcal N(\mathbf f(\mathbf w,\mathbf u_j),\mathbf I)(\boldsymbol\mu_K-\mathbf x)d\mathbf x.
\end{align*}
To find the extreme point, we let the derivative equal to zero.
\begin{align*}
\int^{+\infty}_{-\infty}\frac1K\sum^K_{j=1}\mathcal N(\mathbf f(\mathbf w,\mathbf u_j),\mathbf I)(\boldsymbol\mu_K-\mathbf x)d\mathbf x&=0,\\
\boldsymbol\mu_K-\int^{+\infty}_{-\infty}\frac1K\sum^K_{j=1}\mathcal N(\mathbf f(\mathbf w,\mathbf u_j),\mathbf I)\mathbf xd\mathbf x&=0,\\
\boldsymbol\mu_K=\frac1K\sum^{K}_{j=1}\mathbf f(\mathbf w,\mathbf u_j).
\end{align*}
Now back to derivative of $\mathcal{KL}$ divergence, if $\boldsymbol\mu_K>\frac1K\overset{K}{\underset{j=1}{\sum}}\mathbf f(\mathbf w,\mathbf u_j)$, then $\frac{d}{d\mu_K}\mathcal{KL}[M_K(\mathbf x|\mathbf w)\parallel P_K(\mathbf x|\mathbf w)]>0$. If $\boldsymbol\mu_K<\frac1K\overset{K}{\underset{j=1}{\sum}}\mathbf f(\mathbf w,\mathbf u_j)$, then $\frac{d}{d\mu_K}\mathcal{KL}[M_K(\mathbf x|\mathbf w)\parallel P_K(\mathbf x|\mathbf w)]<0$. Thus $\boldsymbol\mu_K=\frac1K\overset{K}{\underset{j=1}{\sum}}\mathbf f(\mathbf w,\mathbf u_j)$ is global minimum point.
\end{proof}

\begin{proof}[\textbf{proof of Lemma 1}]
Let $P(\mathbf x|\mathbf w,\mathbf u)=\mathcal N(\mathbf f(\mathbf w,\mathbf u),\mathbf I)=\underset{k}{\prod} P_k(x|\mathbf w,\mathbf u)$ where $P_k(x|\mathbf w,\mathbf u)=\mathcal N(f_k(\mathbf w,\mathbf u),1)$ is the distribution of $k$th element of $\mathbf x$ and $f_k(\mathbf w,\mathbf u)$ is the $k$th element of $\mathbf f(\mathbf w,\mathbf u)$.

Since $\Vert\mathbf f(\mathbf w,\mathbf u)\Vert<C$, then for all $i$, $|f_k(\mathbf w,\mathbf u)|<C$. Although the means of every latent variable are different, the maximums of $P_k$ are all the same i.e. $\frac1{\sqrt{(2\pi)}}$. We then define
\begin{align*}
R(x)=\left\{
\begin{aligned}
\frac{1}{\sqrt{2\pi}}\exp\left(-\frac{(x-C)^2}{2}\right),\ &x\in (C,+\infty); \\
\frac{1}{\sqrt{2\pi}}\qquad ,\ &x\in[-C,+C];\\
\frac{1}{\sqrt{2\pi}}\exp\left(-\frac{(x+C)^2}{2}\right),\ &x\in (-\infty,-C). 
\end{aligned}
\right.
\end{align*}
Apparently, $R(x)$ is continuous on $(-\infty,+\infty)$ and 
\begin{align*}
\int^{+\infty}_{-\infty} R(x)dx&=\int^{+\infty}_C \frac{1}{\sqrt{2\pi}}\exp\left(-\frac{(x-C)^2}{2}\right)dx+\int^C_{-C}\frac{1}{\sqrt{2\pi}}dx\\
&+\int^{-C}_{-\infty}\frac{1}{\sqrt{2\pi}}\exp\left(-\frac{(x+C)^2}{2}\right)dx=\frac12+\frac{2C}{\sqrt{2\pi}}+\frac12\\
&=1+\frac{2C}{\sqrt{2\pi}}<+\infty.
\end{align*}
Thus $R(x)$ is integrable on $(-\infty,+\infty)$. Now we prove $P_k(x|\mathbf w,\mathbf u,1)\leq R(x)$ for any $x$, $k$, $\mathbf w$, $\mathbf u$.
It is trivial that $P_k(x|\mathbf w,\mathbf u)\leq \frac1{\sqrt{2\pi}}=R(x)$ when $x\in[-C,+C]$.When $x\in(C,+\infty)$,
\begin{align*}
P_k(x|\mathbf w,\mathbf u)&=\frac{1}{\sqrt{2\pi}}\exp\left(-\frac{(x-f_k(\mathbf w,\mathbf u))^2}{2}\right)\\
&\leq \frac{1}{\sqrt{2\pi}}\exp\left(-\frac{(x-C)^2}{2}\right)\\
&=R(x).
\end{align*}
The second line of equation above depends on $|f_k(\mathbf w,\mathbf u)|<C$ and $0<x-C<x-f_k(\mathbf w,\mathbf u)$ when $x\in(C,+\infty)$.
The condition that $x\in(-\infty,-C)$ is similar. Since $x-f_k(\mathbf w,\mathbf u)\leq x-C<0$,
\begin{align*}
P_k(x|\mathbf w,\mathbf u)&=\frac{1}{\sqrt{2\pi}}\exp\left(-\frac{(x-f_k(\mathbf w,\mathbf u))^2}{2}\right)\\
&\leq \frac{1}{\sqrt{2\pi}}\exp\left(-\frac{(x+C)^2}{2}\right)\\
&=R(x).
\end{align*}
Then $P_k(x|\mathbf w,\mathbf u)\leq R(x)$ for all $x\in(-\infty,+\infty)$. Let $\mathbf R(\mathbf x)=\overset{\dim\mathbf x}{\underset{k=1}{\prod}}R(x_k)$, thus $P(\mathbf x|\mathbf w,\mathbf u)=\prod P_k(x|\mathbf w,\mathbf u)\leq \mathbf R(\mathbf x)$. Then for any $K$, 
\begin{align*}
M_K(\mathbf x|\mathbf w)=\frac1K \sum^{K}_{j=1} P(\mathbf x|\mathbf w,\mathbf u_j)\leq \frac1K \sum^K_{j=1} \mathbf R(\mathbf x)=\mathbf R(\mathbf x).
\end{align*}
Then for any non-negative integrable function $g(y,\mathbf x)$ which is monotone related to $y$, $g\left(M_K(\mathbf x|\mathbf w),\mathbf x\right)\leq g\left(\mathbf R(\mathbf x),\mathbf x\right)$. By the dominated convergence theorem,
\begin{align*}
\lim_{K\to\infty}\int^{+\infty}_{-\infty} g\left(M_K(\mathbf x|\mathbf w),\mathbf x\right)d\mathbf x=\int^{+\infty}_{-\infty}\lim_{K\to\infty} g\left(M_K(\mathbf x|\mathbf w),\mathbf x\right)d\mathbf x.
\end{align*}
\end{proof}

\begin{proof}[\textbf{proof of Theorem 2}]
According to the proof in \textbf{Lemma 1}, let $g(y,x)=|x|y$, thus $g\left(M_K(\mathbf x|\mathbf w),x_i\right)\leq g\left(\mathbf R(\mathbf x),x_i\right)$. Then for every index of element in $\mathbf x$, 
\begin{align*}
\bigg|\int^{+\infty}_{-\infty}x_i M_K(\mathbf x|\mathbf w)d\mathbf x\bigg|&\leq \int^{+\infty}_{-\infty}\big|x_i M_K(\mathbf x|\mathbf w)\big|d\mathbf x\\
&\leq \int^{+\infty}_{-\infty} \big|x_i\mathbf R(\mathbf x)\big|d\mathbf x.
\end{align*}
By the definition of $\mathbf R(\mathbf x)$, it is easy to know $\int^{+\infty}_{-\infty} \big|x_i\mathbf R(\mathbf x)\big|d\mathbf x<+\infty$. Then by the dominated convergence theorem,

\begin{align*}
\int^{+\infty}_{-\infty} x_i P(\mathbf x|\mathbf w)d\mathbf x&=\int^{+\infty}_{-\infty}\lim_{K\to\infty} x_i M_K(\mathbf x|\mathbf w)d\mathbf x\\
&=\lim_{K\to\infty}\int^{+\infty}_{-\infty}x_i M_K(\mathbf x|\mathbf w)d\mathbf x\\
&=\lim_{K\to\infty}\left(\frac1K \sum^{K}_{j=1}f_i(\mathbf w,\mathbf u_j)\right)\\
&=\lim_{K\to\infty}\left(\boldsymbol\mu_K\right)_i.
\end{align*} 
So $E(\boldsymbol\xi)=\underset{K\to\infty}{\lim}\boldsymbol\mu_K$. Similarily, let $h(y,x_i)=(\mathbf x)_i^2y$, we have $x_i^2 M_K(\mathbf x|\mathbf w)\leq x_i^2\mathbf R(\mathbf x)$ for all $K$ and $i$. Then by the dominated convergence theorem,
\begin{align*}
\left((E(\boldsymbol\xi))^2+D(\boldsymbol\xi)\right)_i&=\int^{+\infty}_{-\infty} x_i^2 P(\mathbf x|\mathbf w)d\mathbf x\\
&=\int^{+\infty}_{-\infty}\lim_{K\to\infty} x_i^2 M_K(\mathbf x|\mathbf w)d\mathbf x\\
&=\lim_{K\to\infty}\int^{+\infty}_{-\infty} x_i^2 M_K(\mathbf x|\mathbf w)d\mathbf x\\
&=\lim_{K\to\infty}\left((\boldsymbol\mu_K)^2+\mathbf 1\right)_i.
\end{align*}
Then $D(\boldsymbol\xi)=1$, this completes the proof.
\end{proof}

\begin{proof}[\textbf{proof of Theorem 3}]
By the \textbf{Theoerm 2}, we have $P(\mathbf x|\mathbf w)=\mathcal N\left(\underset{K\to\infty}{\lim}\boldsymbol\mu_K,\mathbf I\right)$. So for any $\mathbf x$, $\mathbf w$ and $\mathbf u$, distribution $P_K(\mathbf x|\mathbf w)$ converges to $P(\mathbf x|\mathbf w)$ point by point. Then by the definition of convergence in distribution, $\boldsymbol\eta_K$ converges to $\boldsymbol\xi$ in disribution.

According to the \textbf{Theorem 2}, mean and covariance matrix of $P_K(\mathbf x|\mathbf w)$  which is independent with $\mathbf x$ is convergence. So we can exchange the limit and integration operation on $h(P_K(\mathbf x|\mathbf w))$ for any funcion $h$ with respect to $\mathbf x$, then by the definition of $\mathcal{KL}$ divergence, 
\begin{align*}
&\lim_{K\to\infty}\mathcal{KL}[P_K(\mathbf x|\mathbf w)\parallel P(\mathbf x|\mathbf w)]\\
=&\lim_{K\to\infty}\int^{+\infty}_{-\infty} P_K(\mathbf x|\mathbf w)\log\left(\frac{P_K(\mathbf x|\mathbf w)}{P(\mathbf x|\mathbf w)}\right)d\mathbf x\\
=&\lim_{K\to\infty}\int^{+\infty}_{-\infty} \mathcal N(\boldsymbol\mu_K,\mathbf I)\log\left(\frac{\mathcal N\left(\boldsymbol\mu_K,\mathbf I\right)}{P(\mathbf x|\mathbf w)}\right)d\mathbf x\\
=&\int^{+\infty}_{-\infty} \mathcal N\left(\lim_{K\to\infty}\boldsymbol\mu_K,\mathbf I\right)\log\left(\frac{\mathcal N\left(\underset{K\to\infty}{\lim}\boldsymbol\mu_K,\mathbf I\right)}{P(\mathbf x|\mathbf w)}\right)d\mathbf x\\
=&\mathcal{KL}[P(\mathbf x|\mathbf w)\parallel P(\mathbf x|\mathbf w)]=0.
\end{align*}
The proof of equation $\underset{K\to\infty}{\lim}\mathcal{KL}[P(\mathbf x|\mathbf w)\parallel P_K(\mathbf x|\mathbf w)]=0$ is similar.
\end{proof}

%\begin{proof}[\textbf{proof of Theorem 4}]
%For $K=1$, the first part of ELBO $\log P_K(\mathbf x|\mathbf w)=-\frac12(\mathbf x-f(\mathbf w,\mathbf u_1))^2$. By the weak convergence approximation algorithms, $\mathbf u_1\sim Q(\mathbf u|\mathbf x)$ is sample after sampling $\mathbf z$ and cover the value of $\mathbf u$ in $\mathbf z$, then $\mathbf w\sim Q(\mathbf w|\mathbf x)$. Thus, the unbiased estimation of $\log P_K(\mathbf x|\mathbf w)$ when $K=1$ is 
%\begin{align*}
%E_{Q(\mathbf w|\mathbf x)}\left(E_{Q(\mathbf u|\mathbf x)}\log P_K(\mathbf x|\mathbf w)\right)&=\int^{+\infty}_{-\infty} Q(\mathbf w|\mathbf x) E_{Q(\mathbf u_1|\mathbf x)}\log P_K(\mathbf x|\mathbf w,\mathbf u_1)d\mathbf w\\
%&=\int^{+\infty}_{-\infty} Q(\mathbf z|\mathbf x) E_{Q(\mathbf u|\mathbf x)}\log P(\mathbf x|\mathbf w,\mathbf u)d\mathbf wd\mathbf u\\
%&=E_{Q(\mathbf z|\mathbf x)}\left(E_{Q(\mathbf u|\mathbf x)}\log P(\mathbf x|\mathbf z)\right).
%\end{align*}
%However $E_{Q(\mathbf w|\mathbf x)}\log P(\mathbf x|\mathbf w)=E_{Q(\mathbf w|\mathbf x)}\left(\log E_{Q(\mathbf u|\mathbf w,\mathbf x)}P(\mathbf x|\mathbf z)\right)$ and $\log$ is strict concave function, then
%\begin{align*}
%E_{Q(\mathbf z|\mathbf x)}\left(E_{Q(\mathbf u|\mathbf x)}\log P(\mathbf x|\mathbf z)\right)&<E_{Q(\mathbf w|\mathbf x)}\left(\log E_{Q(\mathbf u|\mathbf w,\mathbf x)}P(\mathbf x|\mathbf z)\right)\\
%&=E_{Q(\mathbf w|\mathbf x)}\left(\log E_{Q(\mathbf u|\mathbf x)}P(\mathbf x|\mathbf z)\right).
%\end{align*}
%The last step depends on the independence of $\mathbf w$ and $\mathbf u$. This completes the proof.
%\end{proof}

\begin{proof}[\textbf{proof of Theorem 4}]
First of all, we need to turn the question in integration into the question in norm. Assume $\bigg\Vert\frac1K\underset{j=1}{\overset{K}{\sum}}f(\mathbf w,\mathbf u_j)-E(\boldsymbol\xi)\bigg\Vert\leq \frac\epsilon2$, we have
\begin{align*}
&\big|E_{Q(\mathbf w|\mathbf x)}\log P_K(\mathbf x|\mathbf w)-E_{Q(\mathbf w|\mathbf x)}\log P(\mathbf x|\mathbf w)\big|\\
=&\bigg|\int^{+\infty}_{-\infty} Q(\mathbf w|\mathbf x)\frac12\left(\frac1K\sum^{K}_{j=1}f(\mathbf w,\mathbf u_j)-E(\boldsymbol\xi)\right)^\top\left(2\mathbf x+\frac1K\sum^{K}_{j=1}f(\mathbf w,\mathbf u_j)+E(\boldsymbol\xi)\right)d\mathbf x\bigg|\\
\leq &\int^{+\infty}_{-\infty} 2Q(\mathbf w|\mathbf x)\bigg\Vert\frac1K\underset{j=1}{\overset{K}{\sum}}f(\mathbf w,\mathbf u_j)-E(\boldsymbol\xi)\bigg\Vert d\mathbf x\leq \epsilon.
\end{align*}
According to the definition $\Vert f(\mathbf w,\mathbf u)\Vert\leq 1$ and \textbf{Theorem 2}, we have $\Vert E(\boldsymbol\xi)\Vert\leq 1$. Note that $\mathbf x$ is the vector of images, thus $\Vert \mathbf x\Vert \leq 1$. Then the step three uses the fact that $\bigg\Vert\frac1K\underset{j=1}{\overset{K}{\sum}}f(\mathbf w,\mathbf u_j)\bigg\Vert\leq 1$, $\Vert E(\boldsymbol)\Vert\leq1$ and $\Vert\mathbf x\Vert\leq1$. Now we only need to achieve $\bigg\Vert\frac1K\underset{j=1}{\overset{K}{\sum}}f(\mathbf w,\mathbf u_j)-E(\boldsymbol\xi)\bigg\Vert\leq \frac\epsilon2$.
When $\dim\mathbf u$=1, by the definition of $K_c$, when $K\geq K_c$, we have $\bigg\Vert\frac1K\underset{j=1}{\overset{K}{\sum}}f(\mathbf w,\mathbf u_j)-E(\boldsymbol\xi)\bigg\Vert\leq\frac\epsilon6\leq \frac\epsilon2$. 

We only prove the condition of $\dim\mathbf u=2$, the proof of condition that $\mathbf u$ in higher dimension is similar. Let $\dim\mathbf u=\dim\mathbf v=1$. Since the sequence of $\mathbf u_j$ and $\mathbf v_j$ are sampled from $Q(\mathbf w,\mathbf v|\mathbf x)$, without loss of generality, we can assume $\mathbf u_1,...,\mathbf u_{K_1}$, $\mathbf v_1,...,\mathbf v_{K_2}$. Then we have 

\begin{align*}
&\bigg\Vert\frac1{K_1K_2}\sum^{K_1}_{i=1}\sum^{K_2}_{j=1} f(\mathbf w,\mathbf u_i,\mathbf v_j)-E(\boldsymbol\xi)\bigg\Vert\\
\leq &\bigg\Vert \frac1{K_1K_2}\sum^{K_1}_{i=1}\sum^{K_2}_{j=1} f(\mathbf w,\mathbf u_i,\mathbf v_j)-\frac1K_1\sum^{K_1}_{i=1} f(\mathbf w,\mathbf u_i)\bigg\Vert+\bigg\Vert \frac1{K_1}\sum^{K_1}_{i=1} f(\mathbf w,\mathbf u_i)-E(\boldsymbol\xi)\bigg\Vert\\
\leq & \frac1{K_1}\sum^{K_1}_{i=1}\bigg\Vert \frac1{K_2}\sum^{K_2}_{j=1}f(\mathbf w,\mathbf u_i,\mathbf v_j)-f(\mathbf w,\mathbf v_j)\bigg\Vert+\bigg\Vert \frac1{K_1}\sum^{K_1}_{i=1}f(\mathbf u_i)-E(\boldsymbol\xi)\bigg\Vert.
\end{align*}

By the definition of $K_c$, when $K_1\geq K_c$ and $K_2\geq K_c$, 
\begin{align*}
\bigg\Vert\frac1{K_1K_2}\sum^{K_1}_{i=1}\sum^{K_2}_{j=1} f(\mathbf w,\mathbf u_i,\mathbf v_j)-E(\boldsymbol\xi)\bigg\Vert\leq \frac\epsilon2+\frac\epsilon2=\epsilon.
\end{align*}
Thus $K=K_1K_2\leq K_c^2$.
\end{proof}

\begin{proof}[\textbf{proof of Theorem 5}]
By the expression of conditional partitioned Gaussian distribution, we know $\mathbf u|\mathbf w,\mathbf x\sim \mathcal N\left(\boldsymbol\mu_{\mathbf u|\mathbf w},\boldsymbol\Sigma_{\mathbf u|\mathbf w}\right)$, where 
\begin{align*}
\boldsymbol\mu_{\mathbf u|\mathbf w}=\boldsymbol\mu_\mathbf u-(\mathbf w-\boldsymbol\mu_{\mathbf w})\boldsymbol\Lambda_{\mathbf w\mathbf u}\boldsymbol\Sigma_{\mathbf u|\mathbf w},\qquad \boldsymbol\Sigma_{\mathbf u|\mathbf w}=\boldsymbol\Lambda_{\mathbf u\mathbf u}^{-1}.
\end{align*}
Then we compute the $\mathcal{KL}$ divergence,
\begin{align*}
&\mathcal{KL}[Q(\mathbf u|\mathbf w,\mathbf x)\parallel P(\mathbf u)]\\
=&\int^{+\infty}_{-\infty} Q(\mathbf u|\mathbf w,\mathbf x)\log\frac{Q(\mathbf u|\mathbf w,\mathbf x)}{P(\mathbf u)}d\mathbf u\\
=&\int^{+\infty}_{-\infty} Q(\mathbf u|\mathbf w,\mathbf x)\log\frac{\frac{1}{(2\pi)^{\frac{\dim\mathbf u}{2}}\det\big|\boldsymbol\Sigma_{\mathbf u|\mathbf w}\big|}\exp(-\frac12\left((\mathbf u-\boldsymbol\mu_{\mathbf u|\mathbf w})\boldsymbol\Sigma_{\mathbf u|\mathbf w}^{-1}(\mathbf u-\boldsymbol\mu_{\mathbf u|\mathbf w})^\top\right)}{\frac{1}{(2\pi)^{\frac{\dim\mathbf u}{2}}}\exp\left(-\frac12\mathbf u\mathbf u^\top\right)}d\mathbf u\\
=&\int^{+\infty}_{-\infty} Q(\mathbf u|\mathbf w,\mathbf x)\frac12\left(\mathbf u\mathbf u^\top-(\mathbf u-\boldsymbol\mu_{\mathbf u|\mathbf w})\boldsymbol\Sigma_{\mathbf u|\mathbf w}^{-1}(\mathbf u-\boldsymbol\mu_{\mathbf u|\mathbf w})^\top\right)-\log\det\big|\boldsymbol\Sigma_{\mathbf u|\mathbf w}\big|d\mathbf u\\.
\end{align*}
Note that $(\mathbf u-\boldsymbol\mu_{\mathbf u|\mathbf w})\boldsymbol\Sigma_{\mathbf u|\mathbf w}^{-\frac12}\sim \mathcal N(0,\mathbf I)$, we have
\begin{align*}
&\mathcal{KL}[Q(\mathbf u|\mathbf w,\mathbf x)\parallel P(\mathbf u)]\\
=&\frac12\left(\int^{+\infty}_{-\infty} Q(\mathbf u|\mathbf w,\mathbf x)\mathbf u\mathbf u^\top d\mathbf u-\dim\mathbf u-\log\det\big|\boldsymbol\Sigma_{\mathbf u|\mathbf w}\big|\right)\\
=&\frac12\left(\sum^{\dim\mathbf u}_{i=1}\left(E(u_i|\mathbf w,\mathbf x)\right)^2+D(u_i|\mathbf w,\mathbf x)-\dim\mathbf u-\log\det|\boldsymbol\Sigma_{\mathbf u|\mathbf w}\big|\right)\\
=&\frac12\sum^{\dim\mathbf u}_{i=1}\left((\boldsymbol\mu_{\mathbf u|\mathbf w})_i+(\boldsymbol\Sigma_{\mathbf u|\mathbf w})_{ii}-1\right)-\frac12\log\det\big|\boldsymbol\Sigma_{\mathbf w\mathbf w}\big|.
\end{align*}
\end{proof}

\begin{proof}[\textbf{proof of Theorem 6}]
Firstly, we have the equation,
\begin{align*}
&E_{Q(\mathbf w|\mathbf x)}\left(\mathcal{KL}[Q(\mathbf u|\mathbf w,\mathbf x)\parallel P(\mathbf u)]\right)+E_{Q(\mathbf u|\mathbf x)}\left(\mathcal{KL}[Q(\mathbf w|\mathbf u,\mathbf x)\parallel P(\mathbf w)]\right)\\
&-\mathcal{KL}[Q(\mathbf z|\mathbf x)\parallel Q(\mathbf w|\mathbf x)Q(\mathbf u|\mathbf x)]\\
=&\int^{+\infty}_{-\infty} Q(\mathbf z|\mathbf x)\log\frac{Q(\mathbf u|\mathbf w,\mathbf x)Q(\mathbf w|\mathbf u,\mathbf x)Q(\mathbf w|\mathbf x)Q(\mathbf u|\mathbf x)}{P(\mathbf z)Q(\mathbf z|\mathbf x)}\\
=&\mathcal{KL}[Q(\mathbf z|\mathbf x)\parallel P(\mathbf z)].
\end{align*}
Thus
\begin{align*}
&E_{Q(\mathbf w|\mathbf x)}\left(\mathcal{KL}[Q(\mathbf u|\mathbf w,\mathbf x)\parallel P(\mathbf u)]\right)-MI\\
=&\mathcal{KL}[Q(\mathbf z|\mathbf x)\parallel P(\mathbf z)]-E_{Q(\mathbf u|\mathbf x)}\left(\mathcal{KL}[Q(\mathbf w|\mathbf u,\mathbf x)\parallel P(\mathbf w)]\right)\\
\leq &\mathcal{KL}[Q(\mathbf z|\mathbf x)\parallel P(\mathbf z)],
\end{align*}
\begin{align*}
&E_{Q(\mathbf w|\mathbf x)}\left(\mathcal{KL}[Q(\mathbf u|\mathbf w,\mathbf x)\parallel P(\mathbf u)]\right)-MI\\
=&\mathcal{KL}[Q(\mathbf z|\mathbf x)\parallel P(\mathbf z)]-E_{Q(\mathbf u|\mathbf x)}\left(\mathcal{KL}[Q(\mathbf w|\mathbf u,\mathbf x)\parallel P(\mathbf w)]\right)\\
=&E_{Q(\mathbf w|\mathbf u,\mathbf x)}\left(\mathcal{KL}[Q(\mathbf u|\mathbf x)\parallel P(\mathbf u)]\right)\geq0.
\end{align*}
It should be pointed out that $\mathcal{KL}[Q(\mathbf z|\mathbf x)\parallel P(\mathbf z)]$ is the constant with respect to the selection of $\mathbf w$ and $\mathbf u$. So maximize the $E_{Q(\mathbf w|\mathbf x)}\left(\mathcal{KL}[Q(\mathbf u|\mathbf w,\mathbf x)\parallel P(\mathbf u)]\right)-MI$ is equivalent to minimize the $E_{Q(\mathbf u|\mathbf x)}\left(\mathcal{KL}[Q(\mathbf w|\mathbf u,\mathbf x)\parallel P(\mathbf w)]\right)$
\end{proof}